\documentclass[journal,twoside]{IEEEtran}

\usepackage{amsmath}
\usepackage{amssymb}
\usepackage{graphicx}

\usepackage{bm}
\usepackage[mathscr]{eucal}
\allowdisplaybreaks[4]
\newcommand{\vertiii}[1]{\vert\kern-0.25ex\vert\kern-0.25ex\vert #1}

\usepackage{color}
\usepackage{multirow}
\usepackage{makecell} % cell newline

\makeatletter
\renewcommand{\maketag@@@}[1]{\hbox{\m@th\normalsize\normalfont#1}} % normal eq. label
\makeatother

\begin{document}

\title{Design and Analysis of Robust Resilient Diffusion over Multi-Task Networks Against Byzantine Attacks}

\author{Tao Yu,~\IEEEmembership{Member,~IEEE},
        Rodrigo C. de Lamare,~\IEEEmembership{Senior Member,~IEEE},
        and Yi Yu,~\IEEEmembership{Member,~IEEE}
\thanks{Manuscript received June 9, 2021. This work was supported in part by the Natural Science Foundation of Sichuan, China, in 2022, the National Natural Science Foundation of China under Grant 61901400, and the Scientific Research Starting Project of SWPU under Grant 2019QHZ015. {\it (Corresponding author: Tao Yu.)}}
\thanks{Tao Yu is with the School of Electrical Engineering and Information, Southwest Petroleum University, Chengdu 610500, P. R. China (e-mail: yutao@swpu.edu.cn).}
\thanks{Rodrigo C. de Lamare is with the Centre for Telecommunications Studies, Pontifical Catholic University of Rio de Janeiro, Rio de Janeiro 22451-900, Brazil, and also with the Department of Electronic Engineering, University of York, York YO10 5DD, U.K. (e-mail: delamare@puc-rio.br).}
\thanks{Yi Yu is with the School of Information Engineering, Robot Technology Used for Special Environment Key Laboratory of Sichuan Province, Southwest University of Science and Technology, Mianyang 621010, P. R. China (e-mail: yuyi\_xyuan@163.com).}
\vspace{-20pt}
}

%\markboth{IEEE Transactions on Signal Processing}
%{YU \MakeLowercase{\textit{et al.}}: Robust Resilient Diffusion over Multi-Task Networks Against Byzantine Attacks}

\maketitle

\begin{abstract}
This paper studies distributed diffusion adaptation over clustered multi-task networks in the presence of impulsive interferences and Byzantine attacks. We develop a robust resilient diffusion least mean Geman-McClure-estimation (RDLMG) algorithm based on the cost function used by the Geman-McClure estimator, which can reduce the sensitivity to large outliers and make the algorithm robust under impulsive interferences. Moreover, the mean sub-sequence reduced method, in which each node discards the extreme value information of cost contributions received from its neighbors, can make the network resilient against Byzantine attacks. In this regard, the proposed RDLMG algorithm ensures that all normal nodes converge to their ideal states with cooperation among nodes. A statistical analysis of the RDLMG algorithm is also carried out in terms of mean and mean-square performances. Numerical results evaluate the proposed RDLMG algorithm in applications to multi-target localization and multi-task spectrum sensing.
\end{abstract}


\section{Introduction}

\IEEEPARstart{D}{istributed} diffusion techniques involving networks
of agents have been broadly investigated in the last decade, where
the interactive networks among nodes with cooperative relations are
generally modeled by non-negative graphs \cite{Sayed2014Adaptation}.
Distributed cooperation could enable adaptation and learning
strategies through local information exchange among interactive
nodes. Considering that the nodes cooperate with their neighboring
nodes to process streaming data, the diffusion strategies commonly
perform an adaptation step where a node adaptively learns from its
own information, and a combination step where the node aggregates
the information collected from its neighbors. Other nodes in the
network perform similar steps as well, which is known as the
adapt-then-combine (ATC) mechanism \cite{Sayed2014Adaptive}.
Diffusion strategies with the ATC mechanism have proved to be quite
effective in solving distributed optimization problems in a large
number of applications, such as target localization and tracking
\cite{Chen2012Diffusion,FernandezBes2017Adaptive}, collaborative
spectral estimation
\cite{DiLorenzo2013Distributed,Miller2016Distributed}, intrusion
detection and clustering
\cite{Sayed2013Diffusion,Zhao2015Distributed}, and multi-robot
systems \cite{Knorn2016Overview,Yuan2019Exact}. Two fundamental
problems for distributed diffusion are formulated as single-task and
multi-task problems. The single-task one means that all nodes
achieve a common objective, whereas in multi-task scenarios the
nodes are divided into groups and the nodes within the identical
group reach the same objective \cite{Nassif2020Multitask}.
Applications of multi-task strategies, where the objectives of
different nodes or grouped nodes will be distinct, have been related
to multi-target localization and tracking
\cite{Chen2014Multitask,Nassif2016Multitask}, cooperative spectrum
sensing \cite{Chen2017Multitask,PlataChaves2017Heterogeneous},
node-specific speech enhancement
\cite{Hassani2017Multitask,Gogineni2020Improving}, and web page
categorization and ranking \cite{Chen2015Diffusion}.

The conventional diffusion least mean-square (DLMS) algorithm has
been applied to multi-task scenarios in which the nodes cooperate
with each other to estimate distinct parameter vectors using
stochastic gradient descent (SGD) techniques
\cite{Chen2014Multitask,intadap,jio,jidf,dce,saalt,jiodf,als,doaalrd,aaidd,lrcc,dynovs,dqalms,msgamp,dqarls,bimsgamp,rapa}.
Later on, theoretical analysis of DLMS over multi-task networks has
been carried out, where the mean behavior and the networked
mean-square deviation (MSD) were discussed \cite{Chen2015Diffusion}.
Following this research strand, several variants of the DLMS
algorithm were proposed over multi-task networks. If the network
suffers from some uncertainties such as the topology change and link
failure, then the nodes may not execute synchronously, and
asynchronous events could hamper the information exchange among
clusters. The work in \cite{Nassif2016Multitask} introduced an
improved DLMS strategy over asynchronous multi-task networks and
discussed the convergence conditions. A class of DLMS algorithms was
proposed to solve the multi-task estimation problem in
\cite{Nassif2017Diffusion}, whose tasks for neighboring nodes were
related by linear equality constraints. As discussed in
\cite{Wang2017Multitask}, it has been proved that the networked MSD
performance for all nodes relies on the inter-cluster cooperation
over multi-task networks. Utilizing a control variable to maintain
inter-cluster cooperation among neighboring clusters, an affine
projection-based multi-task diffusion strategy was further proposed
to obtain improved convergence performance
\cite{Gogineni2020Improving}.

Generally, the network environments are assumed to be secure, which
means that there is no malicious cyber attacks over networks, and
information exchange among nodes are ideal. In practical
environments, however, large-scale networked systems have many
potential vulnerable points subject to malicious attacks, which are
prone to topology changes or link failures
\cite{Chen2018Internet,Liu2018Secure}. A typical class of malicious
cyber attacks, known as the Byzantine attacks, which can be
completely aware of the networked systems, could ruin the network
via eavesdropping the information and transmitting fictitious
messages \cite{LeBlanc2013Resilient}. Moreover, the nodes suffering
from Byzantine attacks can leverage arbitrarily malicious rules and
send distinct information to different neighbors at the same time.
In terms of attack capabilities, Byzantine attacks involve many
types of malicious behaviors and commonly represent the
``worst-case'' adversarial situation
\cite{Wu2017Secure,Chen2018Resilient,Mitra2019Byzantine}.
Byzantine-resilient decentralized optimization methods have been
developed in \cite{Yang2019ByRDiE} and \cite{Yang2020Adversary}, in
which every regular agent uses its local data samples and applies
coordinate gradient descent to update its local model at every
iteration. The work in \cite{Peng2021Byzantine} introduced a total
variation norm-penalized approximation to handle Byzantine attacks
and proposed a stochastic gradient approach to solve the
Byzantine-resilient decentralized optimization over static and
time-varying networks. Therefore, it is fundamental to design secure
distributed algorithms to enable to resist malicious attacks and to
allow normal nodes still to achieve their ideal objectives. We refer
to such secure distributed algorithms as being resilient to
malicious attacks and such network for resilience to adversaries
\cite{Li2020Resilient}.

To cope with malicious cyber attacks, several secure distributed
strategies have been developed in recent years, mainly towards
improving the distributed SGD solver of the underlying optimization
tasks over networks. A secure DLMS algorithm was proposed in
\cite{Liu2018Secure}, in which this distributed algorithm was
regarded as a mixed strategy consisting of the non-cooperative LMS
(NC-LMS) and the DLMS hybrid mechanisms. Along this line of
research, a correction-based secure DLMS algorithm was proposed in
\cite{Chang2020Correction}, where a correction step was appended
between the adaptation and combination. The aforesaid algorithms
considered false data injection (FDI) attacks that satisfied linear
combinations of the regression vector and the attack vector, and a
DLMS algorithm based on the Kullback-Leibler divergence was further
developed to detect adversaries \cite{Hua2020Secure}. Besides, the
work in \cite{Shi2020Secure} proposed a secure distributed algorithm
relying on attack detection to deal with the multi-task estimation
problem against FDI attacks. Considering Byzantine attacks, a secure
distributed estimation was established in \cite{Chen2018Resilient}
through attack detection raising a flag to indicate the estimate
beyond a threshold, where the node then combined estimates of
neighboring nodes with local information. In order to make the
network resilient against Byzantine attacks, a family of mean
sub-sequence reduced (MSR) algorithms has been investigated, where
each node removes some extreme values received from its neighboring
nodes and updates its state to be the mean of the remaining values
\cite{LeBlanc2013Resilient,Abbas2018Improving}. The majority of the
distributed optimization problems under adversarial nodes consider
consensus-based optimization protocols, in which the node gathers
the states of its neighbors and removes some extreme values that are
larger or smaller than its own state value
\cite{Sundaram2019Distributed}. The work in \cite{Zhao2020Resilient}
exploited the trusted nodes which can not be compromised by
adversarial attacks and formed a connected dominating set in the
original network to constrain effects of adversarial nodes. With the
help of a candidate-removal mechanism, a distributed optimization
problem was addressed to generate reliable states in the presence of
malicious nodes \cite{Lu2021Distributed}. There are numerous
developments of distributed estimation problems in the presence of
Byzantine nodes, where the node removes some extreme values received
from its neighbors and the resilience is achieved. Resilience of
consensus-based distributed estimation has been studied in
\cite{Chen2019aResilient,Chen2019bResilient} by leveraging local
observation and network connectivity. A distributed estimator with
resilience to adversaries was further investigated in
\cite{Chen2020Resilient}, where each node iteratively updated a
local estimate after removing abnormal observations as a weighted
sum of its previous estimate. Recently, an intriguing work in
\cite{Li2020Resilient} developed a resilient diffusion least
mean-square (RDLMS) algorithm motivated by a MSR approach over
multi-task networks in the presence of Byzantine attacks, and also
presented the trade-off between the resilience of the distributed
strategy and its performance of the networked MSD in qualitative
analysis.

The majority of the aforementioned secure distributed algorithms
focus on Gaussian interferences. However, in realistic
circumstances, there are many non-Gaussian interferences which can
be expressed by the impulsive feature, whose probability density
functions have heavier tail than that of Gaussian interferences
\cite{Yu2020Diffusion,Yu2021Robust,Yu2021RobustAdaptive}. Although
the above distributed algorithms using the mean-square error (MSE)
criterion have good performance under Gaussian measurement noises,
they often have significant performance degradation under impulsive
interferences, especially due to the inter-node cooperation
\cite{Chen2018Diffusion,Wilson2020Robust}. To execute well in the
presence of impulsive interferences, several robust algorithms have
been designed to reduce the sensitivity to large outliers.
Typically, the lower-order statistics of the error were used to
suppress impulsive measurement noises, where the criteria involving
the mean absolute error \cite{Ni2021Multitask}, the mean $p$-power
error \cite{Lu2018Performance}, etc. Some M-estimate approaches were
developed based on cost functions such as the Huber functions to
counteract the negative influences of impulsive interferences
\cite{Yu2020Diffusion}. However, such robust cost functions can not
commonly be smooth everywhere. Hence, other smooth robust estimators
were further developed by utilizing the saturation property of error
nonlinearities like the maximum correntropy criterion
\cite{Chen2018Diffusion,Ma2016Diffusion}, and the Geman-McClure (GM)
estimator \cite{Lu2019Recursive}. Nevertheless, it should be
remarked that in scenarios with impulsive interferences distributed
diffusion over multi-task networks against malicious cyber attacks,
despite its paramount importance, is still open to the best of our
knowledge and motivates this work.

In this paper, we develop and analyze a robust resilient diffusion
algorithm over clustered multi-task networks in the presence of
impulsive interferences and Byzantine attacks. The main
contributions are:
\begin{enumerate}
  \item We propose a novel resilient diffusion least mean Geman-McClure-estimation (RDLMG) algorithm based on the GM estimator and the MSR method over multi-task networks, which ensures robustness under impulsive interferences and resilience against Byzantine attacks.
  \item The theoretical performance analyses of the proposed RDLMG algorithm are rigorously derived and analyzed for multi-task estimation in the mean and mean-square senses. In particular, a closed-form expression of the steady-state networked MSD for all normal nodes over the multi-task network can be accurately predicted.
  \item Simulation studies of the proposed RDLMG algorithm applied to multi-target localization and multi-task spectrum sensing demonstrate its superiority over existing techniques.
\end{enumerate}

The rest of this paper is organized as follows. A network model and
some preliminaries of distributed diffusion are introduced in
Section II. The problem formulation and the proposed RDLMG algorithm
for multi-task estimation are presented in Section III. Theoretical
analyses of the RDLMG algorithm in the mean and mean-square senses
are discussed in Section IV and V, respectively. In Section VI, we
apply the RDLMG algorithm to multi-target localization and
multi-task spectrum sensing. Finally, Section VII concludes this
paper.

\emph{Notations:} Notations in this paper are rather standard. At
time instant, the normal font letters with parentheses represent
scalars. The lowercase boldface letters with parentheses represent
column vectors. The uppercase boldface letters with parentheses
represent matrices. Other symbols and notations will be explained in
the corresponding text.

\section{Network Model and Preliminaries}

\subsection{Network Model}

An interactive network with multiple nodes (also known as a
multi-agent system) can be described by a non-negative graph
$\mathcal{G}=\{\mathcal{V},\mathcal{E}\}$, with the finite indices
set of nodes $\mathcal{V}=\{1,2,\dots,P\}$ and the edges set
$\mathcal{E} \subseteq \mathcal{V} \times \mathcal{V}$. An edge
$(j,i)$ is graphically drawn via an arrow with head node $i$ and
tail node $j$, which means that node $i$ receives the information
from node $j$. Throughout this paper, we consider that the graph has
the bidirectional information flow and the arrow is naturally
omitted over graphs. The node $j$ is a neighbor of node $i$ if
$(j,i)\in\mathcal{E}$, and the indices set of neighbors of node $i$
is defined by $\mathcal{N}_i=\{j \mid (j,i)\in\mathcal{E}\}$,
including node $i$ itself.

At every time instant $n$, each node $i$ acquires the desired output signal $d_i(n)$ and the input signal $u_i(n)$ and the $M$-dimensional tapped-delay regression vector defined as $\bm{u}_i(n)=[u_i(n),u_i(n-1),\dots,u_i(n-M+1)]^\mathrm{T} \in \mathbb{R}^M$, which is mathematically described via a linear form
\begin{align}\label{eq:d}
d_i(n) = \bm{u}_i^\mathrm{T}(n) \bm{w}_i^\mathrm{o} + \eta_i(n)
\end{align}
where $\bm{w}_i^\mathrm{o} \in \mathbb{R}^M$ represents the ideal state vector, and $\eta_i(n)$ is an additive noise.

The objective of each node $i$ in the network is to utilize the streaming data $\{d_i(n),\bm{u}_i(n)\}$ to compute an estimate of the state vector $\bm{w}_i^\mathrm{o}$. In the case of clustered multi-task scenarios, each node $i$ needs to determine its own state vector $\bm w_i^\mathrm{o}$. All nodes are classified into clusters, where the nodes within the identical cluster perform the same state vector, and different clustered nodes are interested in estimating different state vectors \cite{Chen2014Multitask,Zhao2015Distributed}.

\subsection{Distributed Adaptive Algorithms}

\subsubsection{DLMS Algorithm}

Consider the distributed optimization problem of the following MSE criterion as the cost function
\begin{align}
\min_{\bm{w}_i} \sum_{i=1}^P \mathbb{E} \{[d_i(n) - \bm{u}_i^\mathrm{T}(n) \bm{w}_i]^2\}
\end{align}
where $\mathbb{E}\{\cdot\}$ represents the expectation operation. When each node $i$ performs its own individual LMS algorithm, the adaptive solution of NC-LMS can be given by
\begin{align}
\bm{w}_i(n+1) = \bm{w}_i(n) + \mu_i e_i(n) \bm{u}_i(n) && \text{(non-cooperation)}
\end{align}
where $\bm{w}_i(n) = [w_{i,1}(n),w_{i,2}(n),\dots,w_{i,M}(n)]^\mathrm{T} \in \mathbb{R}^M$ is the estimate of $\bm{w}_i^\mathrm{o}$ at time $n$, and
\begin{align}\label{eq:e}
e_i(n) = d_i(n) - \bm{u}_i^\mathrm{T}(n) \bm{w}_i(n)
\end{align}
is the error signal, and the constant $\mu_i>0$ denotes the step size. It means that each node $i$ iteratively updates its estimate without cooperation.

Now consider that the nodes cooperate with their neighbors to process streaming data of local information through the diffusion network, to execute the multi-task estimation. A cost-effective diffusion strategy is DLMS with the ATC mechanism \cite{Chen2015Diffusion}, which solves the MSE minimization problem to get the cooperative adaptive solution
\begin{align}
\bm{\psi}_i(n+1) &= \bm{w}_i(n) + \mu_i e_i(n) \bm{u}_i(n) & \text{(adaptation)} \nonumber\\
\bm{w}_i(n+1) &= \sum_{j \in \mathcal{N}_i} a_{j,i}(n+1) \bm{\psi}_j(n+1) & \text{(combination)}
\end{align}
where the intermediate estimate $\bm{\psi}_i(n)\in \mathbb{R}^M$ is shared among neighboring nodes, and the coefficient $a_{j,i}(n)$ denotes the weight in which node $i$ measures the data receiving from node $j$. It represents the reliable measure of edge $(j,i)$ connecting two neighboring nodes, and the assigned weights satisfy the following conditions
\begin{align}
a_{j,i}(n)\geq 0, \sum_{j \in \mathcal{N}_i} a_{j,i}(n)=1, a_{j,i}(n) = 0~\text{if}~j \notin \mathcal{N}_i   .
\end{align}
The connections among nodes over the graph $\mathcal{G}$ are captured by the adjacency matrix $\bm A (n) = [a_{i,j}(n)] \in \mathbb{R}^{P \times P}$, and its transpose matrix is $\bm A^\mathrm{T}(n)=[a_{j,i}(n)]$, where each entry $a_{j,i}(n)>0$ represents the weight associated with edge if $(j,i)\in \mathcal{E}$, otherwise $a_{j,i}(n)=0$. Then, the rows of matrix $\bm A^\mathrm{T}(n)$ adds up to one, i.e., $\bm A^\mathrm{T}(n) \bm 1_P = \bm 1_P$, where $\bm 1_P$ with size $P \times 1$ represents all its entries equal to one.

At every time $n$, the traditional DLMS with the ATC mechanism performs two steps. The first step is that node $i$ learns from the current estimate $\bm{w}_i(n)$ to the intermediate estimate $\bm{\psi}_i(n+1)$. The second step is an aggregation where node $i$ combines all intermediate estimates of its neighbors to yield its successive estimate $\bm{w}_i(n+1)$.

\subsubsection{DLMG Algorithm}

As we all know, the conventional DLMS algorithm acquires the adaptive solution by relying on the MSE cost, but it may degenerate or even diverge under impulsive noise environments \cite{Yu2020Diffusion,AlSayed2017Robust}. To obtain robustness for the multi-task diffusion strategy under impulsive interferences, we consider a GM estimator belonging to a class of robust M-estimators as
\begin{align}
\Upsilon_i(e_i(n)) = \frac{0.5e_i^2(n)}{1 + \lambda e_i^2(n)}
\end{align}
where $\lambda$ is a positive parameter. The GM estimator is a smooth non-convex loss function, which handles outliers well and may be more applicable than convex M-estimators \cite{Lu2019Recursive,Mandanas2017Robust}. The following scale function can be calculated as
\begin{align}
f_i(e_i(n)) = \frac{\partial \Upsilon_i(e_i(n))}{\partial e_i(n)}\cdot \frac{1}{e_i(n)} = \frac{1}{[1 + \lambda e_i^2(n)]^2}   .
\end{align}

Let $J_i(\bm{w}_i)$ denote the cost function inspired by the GM estimator at node $i$, namely,
\begin{align}\label{eq:global}
J_i(\bm{w}_i) = \mathbb{E} \left\{\frac{0.5[d_i(n) - \bm{u}_i^\mathrm{T}(n) \bm{w}_i]^2}{1+\lambda[d_i(n) - \bm{u}_i^\mathrm{T}(n) \bm{w}_i]^2}\right\}   .
\end{align}
Based on the diffusion strategy that every node fuses its own information and the received information from its neighbors,
we thus consider the minimization problem to solve the distributed optimization as follows \cite{Chen2014Multitask,Chen2015Diffusion}:
\begin{align}
&\min_{\bm w_i} J_i^\mathrm{loc}(\bm w_i) \nonumber\\
&J_i^\mathrm{loc}(\bm w_i) = \sum_{j \in \mathcal{N}_i}a_{j,i} J_i(\bm w_i)  .
\end{align}
Taking advantage of the SGD technique to deduce the distributed adaptive algorithm, the ATC mechanism can be iteratively updated as
\begin{align}
\bm{\psi}_i(n+1) &= \bm{w}_i(n) + \mu_i \frac{1}{[1 + \lambda e_i^2(n)]^2} e_i(n) \bm{u}_i(n)\nonumber\\
\bm{w}_i(n+1) &= \sum_{j \in \mathcal{N}_i} a_{j,i}(n+1) \bm{\psi}_j(n+1)   .
\end{align}
This distributed diffusion strategy is called diffusion least mean GM-estimation (DLMG) algorithm with cooperation among nodes. When each node $i$ performs its own learning strategy, the non-cooperative least mean GM-estimation (NC-LMG) algorithm is given by
\begin{align}
\bm{w}_i(n+1) = \bm{w}_i(n) + \mu_i \frac{1}{[1 + \lambda e_i^2(n)]^2} e_i(n) \bm{u}_i(n)   .
\end{align}

\begin{figure}[!ht]
\centering
\includegraphics[width=2.5in]{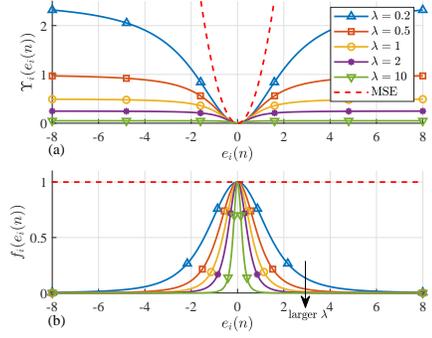}
\caption{(a) Curves of $\Upsilon_i(e_i(n))$ with different $\lambda$. (b) Curves of $f_i(e_i(n))$ with different $\lambda=0.2,0.5,1,2,10$.}
\label{fig:cost_function}
\end{figure}

The curves of $\Upsilon_i(e_i(n))$ and $f_i(e_i(n))$ with different $\lambda$ are shown in Fig. \ref{fig:cost_function}. Whenever the impulsive interferences occur, $e_i(n)$ of each node $i$ becomes quite large, and thus the scale function $f_i(e_i(n))$ reduces to zero, which will handle large outliers well. However, the scale function of MSE are equal to one which can not reduce the sensitivity to large outliers. Besides, it shows that the larger $\lambda$ can result in increased robustness. It also shows that the smaller $\lambda$ can cause the larger magnitude of $f_i(e_i(n))$ indicating the faster convergence around the small $e_i(n)$, but it will increase the system error. So the appropriate $\lambda$ should be selected for ensuring robustness and good performance.

Several fixed combination weights for a single-task are adopted from the degree of each node according to the network topology, such as the Laplacian and Metropolis rules \cite{Sayed2014Adaptation,Sayed2014Adaptive,Sayed2013Diffusion}. However, these fixed weights may not be applicable in multi-task scenarios, where any node performing a distinct state will hamper its neighbors estimating themselves other states of interest. Hence, a class of adaptive weights rules has been introduced for multi-task estimation \cite{Li2020Resilient,Zhao2012Diffusion} as
\begin{align}
a_{j,i}(n+1) =
\left\{\begin{aligned}
&\frac{\gamma_{j,i}^{-2}(n+1)}{\sum_{\ell \in \mathcal{N}_i} \gamma_{\ell,i}^{-2}(n+1)},  & &  j \in \mathcal{N}_i\\
&0, & &  \text{otherwise}
\end{aligned}\right.
\end{align}
where
\begin{align}\label{eq:update_weights}
\gamma_{j,i}^2(n+1) = (1-\nu_i)\gamma_{j,i}^2(n) + \nu_i\|\bm{\psi}_j(n+1) - \bm{w}_i(n)\|_2^2
\end{align}
with $0<\nu_i<1$ being the forgetting factor, and $\|\cdot\|_2$ denotes the $l_2$-norm (or Euclidean norm) of its vector argument.

\emph{Remark 1:} We can also consider other adaptive combination rules as the variations of \eqref{eq:update_weights} in such multi-task scenario. One strategy for adjusting the combination weights in \cite{Chen2015Diffusion} is to use the local one-step approximation for $\bm{w}_i^\mathrm{o}$. Moreover, the adaptive combination rule developed in \cite{Nassif2016Diffusion} contains the smoothing step based on the one-step approximation.

By allowing nodes to adjust their combination weights, this enables nodes to adaptively assign the weights depending on the cooperative contributions from their neighbors. So we do not assume the availability of any prior information beforehand whether the relationships exist among the states of neighboring nodes. During the multi-task estimation, the nodes sensing distinct objective states than their own will gradually assign smaller weights through information exchange. It becomes possible for the nodes to discard their connections during a certain time to neighbors that do not contribute to the same objective. Moreover, the nodes sensing an identical cooperative objective will cluster together to form connected sub-networks \cite{Zhao2015Distributed,Chen2015Diffusion}.

\section{Problem Formulation and Robust Resilient Diffusion Algorithm}

In this section, we formulate the problem and present the proposed robust resilient diffusion algorithm.

\subsection{Byzantine Attack Model}

According to the Byzantine attack characteristics, we consider some malicious nodes over multi-task networks to be adversarial, called Byzantine nodes, and others are normal nodes \cite{Li2020Resilient}. Generally speaking, the Byzantine node represents the ``worst-case'' malicious behavior, which has complete knowledge of the whole system, including the network topology, streaming data and diffusion algorithm employed by normal nodes. In addition, malicious nodes are assumed to be Byzantine nodes in the sense that they can update their states in arbitrary ways, send misinformation to their neighbors, and transmit distinct information to different neighbors \cite{LeBlanc2013Resilient,Wu2017Secure}.

\emph{Attack Objective:} The objective of the Byzantine node in the network is to drive its neighboring normal nodes to reach the malicious states. In detail, we consider the finite indices set of Byzantine nodes $\mathcal{V}^\mathrm{a} \subset \mathcal{V}$, and assume that a Byzantine node $k \in \mathcal{V}^\mathrm{a}$ wants to drive the state of its neighboring normal node $i \in \mathcal{N}_k$ to reach the malicious state $\bm{w}_i^\mathrm{a}$, i.e.,
\begin{align}
\lim_{n \to \infty} \mathbb{E}\{\bm{w}_i(n+1)\} = \bm{w}_i^\mathrm{a}, \forall i \in \mathcal{N}_k, \forall k \in \mathcal{V}^\mathrm{a}   .
\end{align}
Then, the following proposition provides the sufficient condition to reach the attack objective of a Byzantine node $k$ which drives its targeted normal node $i$ by leveraging the local information exchange.

\emph{Proposition 1:} If a sufficiently small value $\mu_i^\mathrm{a}>0$ has been selected such that $\forall j \in \mathcal{N}_i \backslash k$,
\begin{align}\label{eq:transmitting_condition}
\|\mu_i^\mathrm{a}[\bm{w}_i(n)-\bm{w}_i^\mathrm{a}]\|_2 \ll \|\bm{\psi}_j(n+1)-\bm{w}_i(n)\|_2
\end{align}
holds, then the Byzantine node $k$ leverages the gradient-based transmitting message
\begin{align}\label{eq:transmitting_message}
\bm{\psi}_k(n+1) = \bm{w}_i(n) -  \mu_i^\mathrm{a}[\bm{w}_i(n)-\bm{w}_i^\mathrm{a}]
\end{align}
as the intermediate estimate of the targeted node $i$, to finally reach the following attack objective:
\begin{enumerate}
  \item The influence of $\bm{\psi}_j(n+1),\forall j \in \mathcal{N}_i \backslash k$ will be neglected in the presence of the Byzantine node, i.e.,
\begin{align}
a_{j,i}(n+1) \to 0, \forall j \in \mathcal{N}_i \backslash k,~\text{and}~a_{k,i}(n+1) \to 1  .
\end{align}
  \item The state of the targeted normal node will converge to the malicious state
\begin{align}
\mathbb{E}\{\bm{w}_i(n+1)\} \to \bm{w}_i^\mathrm{a}, \forall i \in \mathcal{N}_k, \forall  k \in \mathcal{V}^\mathrm{a}
\end{align}
for a large enough and finite time.
\end{enumerate}

\emph{Proof:} By \eqref{eq:transmitting_message} and the transmitting message $\bm{\psi}_k(n+1)$ as the intermediate estimate of the targeted node $i$, a sufficiently small value of $\mu_i^\mathrm{a}>0$ can ensure the following condition
\begin{align}
\|\bm{\psi}_k(n+1)-\bm{w}_i(n)\|_2 &= \|\mu_i^\mathrm{a}[\bm{w}_i(n)-\bm{w}_i^\mathrm{a}]\|_2 \nonumber\\
&\ll \|\bm{\psi}_j(n+1)-\bm{w}_i(n)\|_2  .
\end{align}
Denote
\begin{align}
\delta_{k,i}(n+1) &\triangleq \|\bm{\psi}_k(n+1)-\bm{w}_i(n)\|_2 \nonumber\\
\delta_{j,i}(n+1) &\triangleq \|\bm{\psi}_j(n+1)-\bm{w}_i(n)\|_2
\end{align}
and suppose the attack starts at time $n+1$, and then the recursive equation at time $n+n^\mathrm{c}$ according to \eqref{eq:update_weights} is calculated as follows:
\begin{align}
&~\gamma_{k,i}^2(n+n^\mathrm{c}) \nonumber\\
=&~(1-\nu_i)\gamma_{k,i}^2(n+n^\mathrm{c}-1) + \nu_i \delta_{k,i}^2(n+n^\mathrm{c}) \nonumber\\
=&~(1-\nu_i)^{n^\mathrm{c}} \gamma_{k,i}^2(n) + \nu_i \sum_{\ell = 0}^{n^\mathrm{c}-1} (1-\nu_i)^\ell \delta_{k,i}^2(n+n^\mathrm{c}-\ell)
\end{align}
and
\begin{align}
&~\gamma_{j,i}^2(n+n^\mathrm{c}) \nonumber\\
=&~(1-\nu_i)\gamma_{j,i}^2(n+n^\mathrm{c}-1) + \nu_i \delta_{j,i}^2(n+n^\mathrm{c}) \nonumber\\
=&~(1-\nu_i)^{n^\mathrm{c}} \gamma_{j,i}^2(n) + \nu_i \sum_{\ell = 0}^{n^\mathrm{c}-1} (1-\nu_i)^\ell \delta_{j,i}^2(n+n^\mathrm{c}-\ell)  .
\end{align}
For large enough $n^\mathrm{c}$, it will have $(1-\nu_i)^{n^\mathrm{c}} \to 0$. In fact, $(1-\nu_i)^{n^\mathrm{c}} \le \epsilon^\mathrm{c}$ with a certain small value $\epsilon^\mathrm{c}$ will converge, and the time required to ensure the convergence is computed as $n^\mathrm{c} \ge \log_{(1-\nu_i)} {\epsilon^\mathrm{c}}$. For a large enough time, since $\delta_{k,i}^2(n+1)\ll\delta_{j,i}^2(n+1)$, it implies that $\gamma_{k,i}^2(n+1)\ll\gamma_{j,i}^2(n+1)$ holds, then it follows that $\gamma_{j,i}^{-2}(n+1)\ll\gamma_{k,i}^{-2}(n+1)$ and $a_{j,i}(n+1) \ll a_{k,i}(n+1)$. From $\sum_{j \in \mathcal{N}_i \backslash k} a_{j,i}(n+1) + a_{k,i}(n+1)=1$, we can obtain $a_{j,i}(n+1) \to 0, \forall j \in \mathcal{N}_i \backslash k$, and $a_{k,i}(n+1) \to 1$. It means that the influence of $\bm{\psi}_j(n+1),\forall j \in \mathcal{N}_i \backslash k$ will be neglected in the presence of the Byzantine node during a finite time.

In addition, for some time $n$ subject to $(1-\nu_i)^n \to 0$, the state of the targeted node $i$ will be attacked as
\begin{align}
\mathbb{E}\{\bm{w}_i(n+1)\} =~& \mathbb{E}\bigg\{\sum_{j \in \mathcal{N}_i \backslash k} a_{j,i}(n+1) \bm{\psi}_j(n+1) \nonumber\\
&+ a_{k,i}(n+1) \bm{\psi}_k(n+1) \bigg\} \nonumber\\
=&~\mathbb{E}\{\bm{\psi}_k(n+1)\} \nonumber\\
=&~\mathbb{E}\{\bm{w}_i(n)\} -  \mu_i^\mathrm{a}[\mathbb{E}\{\bm{w}_i(n)\}-\bm{w}_i^\mathrm{a}] \nonumber\\
=&~(1 -  \mu_i^\mathrm{a})\mathbb{E}\{\bm{w}_i(n)\} + \mu_i^\mathrm{a}\bm{w}_i^\mathrm{a}   .
\end{align}
The recursive vector equation at time $n+n^\mathrm{a}$ is calculated as
\begin{align}\label{eq:attack_weight}
\mathbb{E}\{\bm{w}_i(n+n^\mathrm{a})\} =~& (1-\mu_i^\mathrm{a})^{n^\mathrm{a}} \mathbb{E}\{\bm{w}_i(n)\} \nonumber\\
&+  \sum_{\ell = 0}^{n^\mathrm{a}-1} (1-\mu_i^\mathrm{a})^\ell  \mu_i^\mathrm{a}\bm{w}_i^\mathrm{a}   .
\end{align}
For large enough $n^\mathrm{a}$, it will have $(1-\mu_i^\mathrm{a})^{n^\mathrm{a}} \to 0$. Similarly, the time required to ensure the convergence can be computed as $n^\mathrm{a} \ge \log_{(1-\mu_i^\mathrm{a})} {\epsilon^\mathrm{a}}$ with a small value $\epsilon^\mathrm{a}$. Furthermore, using the sum of geometric series from \eqref{eq:attack_weight} results in $\mathbb{E}\{\bm{w}_i(n+1)\} \to \bm{w}_i^\mathrm{a}$ during a finite time. \hfill $\blacksquare$

\emph{Remark 2:} In practice, as long as a sufficiently small value $\mu_i^\mathrm{a}$ has been taken, the condition \eqref{eq:transmitting_condition} can be guaranteed. The attacker can select a small fixed value $\mu_i^\mathrm{a}$ and observe if the attack succeeds; if not, decrease it to find an appropriate value. Then, the attack objective will be reached during a large enough and finite time.

\subsection{Proposed Robust Resilient Diffusion Algorithm}

It is a fact that if multiple Byzantine nodes transmit the misinformation \eqref{eq:transmitting_message} to the identical targeted node, then the weights will be assigned to sum up to one, and it is not necessary for multiple Byzantine nodes to attack a single normal node. Moreover, the Byzantine node can only drive its direct neighbors and can not impact the nodes beyond its direct neighbors \cite{Li2020Resilient}. Hence, we make the following assumption.

\emph{Assumption 1:} Under the gradient-based Byzantine attack model, we assume that there could be at most one Byzantine node among the neighbors of a normal node.

\emph{Resilient Objective:} We propose a resilient diffusion algorithm for multi-task estimation against Byzantine attacks such that
\begin{align}
\lim_{n \to \infty}\mathbb{E}\{\bm{w}_i(n+1)\} = \bm{w}_i^\mathrm{o}, \forall i \in \mathcal{V} \backslash \mathcal{V}^\mathrm{a}
\end{align}
for all normal nodes, i.e., all normal nodes achieve their ideal states, where the cooperation and information exchange should be utilized among nodes over clustered multi-task networks.

The MSR method, in which each node discards the values received from its neighbors and deviated from its own value, can make the network resilient by updating its state to be the weighted average of the remaining values \cite{LeBlanc2013Resilient}. Inspired by the MSR method, for each node $i \in \mathcal{V} \backslash \mathcal{V}^\mathrm{a}$, the combination step of the diffusion strategy aggregates the intermediate estimates of its neighbors by discarding one extreme value of local information to obtain an accurate estimate. This extreme value will then be removed from each node.

From the Byzantine node's perspective, the attack objective is to maximize the local cost $J_i(\bm{w}_i(n))$ of the targeted normal node, whereas the diffusion strategy aims to obtain the minimization cost. Since $\bm{w}_i(n) = \sum_{j \in \mathcal{N}_i} a_{j,i}(n) \bm{\psi}_j(n)$ and $\sum_{j \in \mathcal{N}_i} a_{j,i}(n)=1$, the local cost $J_i(\bm{w}_i(n)), \forall i \in \mathcal{V} \backslash \mathcal{V}^\mathrm{a}$ can be calculated as
\begin{align}\label{eq:local_cost}
&~J_i(\bm{w}_i(n)) \nonumber\\
=&~J_i\bigg(\sum_{j \in \mathcal{N}_i} a_{j,i}(n) \bm{\psi}_j(n)\bigg)  \nonumber\\
=&~\mathbb{E} \left\{\frac{0.5\Big[\sum\limits_{j \in \mathcal{N}_i} a_{j,i}(n)\Big(d_i(n) - \bm{u}_i^\mathrm{T}(n) \bm{\psi}_j(n)\Big)\Big]^2}{1+\lambda\Big[\sum\limits_{j \in \mathcal{N}_i} a_{j,i}(n)\Big(d_i(n) - \bm{u}_i^\mathrm{T}(n) \bm{\psi}_j(n)\Big)\Big]^2}\right\} \nonumber\\
\approx &~\frac{0.5 \sum\limits_{j \in \mathcal{N}_i} a_{j,i}^2(n) \mathbb{E} \Big\{  [d_i(n) - \bm{u}_i^\mathrm{T}(n) \bm{\psi}_j(n)]^2\Big\} }{1+\lambda\sum\limits_{j \in \mathcal{N}_i} a_{j,i}^2(n) \mathbb{E} \Big\{  [d_i(n) - \bm{u}_i^\mathrm{T}(n) \bm{\psi}_j(n)]^2\Big\} } \nonumber\\
\propto&~\sum\limits_{j \in \mathcal{N}_i} a_{j,i}^2(n) \mathbb{E} \Big\{  [d_i(n) - \bm{u}_i^\mathrm{T}(n) \bm{\psi}_j(n)]^2\Big\} \nonumber\\
\propto&~ \gamma_{j,i}^{-4}(n)Q_i(\bm{\psi}_j(n))
\end{align}
where $Q_i(\bm{\psi}_j(n)) \triangleq \mathbb{E} \{  [d_i(n) - \bm{u}_i^\mathrm{T}(n) \bm{\psi}_j(n)]^2 \}$.

During the iterations, each node performs the diffusion strategy to minimize its cost $J_i(\bm{w}_i(n))$ requiring the local information exchange among its neighboring nodes. According to \eqref{eq:local_cost}, the contribution of the neighboring node $j$ to the node $i$'s cost $J_i(\bm{w}_i(n))$ can be defined as
\begin{align}
c_{j,i}(n) \triangleq \gamma_{j,i}^{-4}(n)Q_i(\bm{\psi}_j(n))  .
\end{align}
By removing the largest cost contribution $c_{j,i}(n)$ received from the node $i$'s neighbors, the removal set $\mathcal{R}_i(n)$ can be taken as
\begin{align}
\mathcal{R}_i(n) = \Big\{j \bigm| \arg \max_{j \in \mathcal{N}_i} c_{j,i}(n) \Big\}  .
\end{align}

At time $n+1$, the node $i$ adjusts its combination weights $a_{j,i}(n+1)$ and updates the state $\bm{w}_i(n+1)$ through its neighboring information exchange, but without using information form the node in $\mathcal{R}_i(n+1)$. Then, the ATC mechanism can be rewritten as
\begin{align}\label{eq:psi}
\bm{\psi}_i(n+1) &= \bm{w}_i(n) + \mu_i \frac{1}{[1 + \lambda e_i^2(n)]^2} e_i(n) \bm{u}_i(n)  \nonumber\\
\bm{w}_i(n+1) &= \sum_{\scriptscriptstyle j \in \mathcal{N}_i \backslash \mathcal{R}_i(n+1)} a_{j,i}(n+1) \bm{\psi}_j(n+1)   .
\end{align}
Without any prior information, the node does not know beforehand whether there is a Byzantine node in its neighborhood. Note that the message from normal neighboring node may be discarded by the removal set $\mathcal{R}_i(n+1)$. The removed messages do not affect the convergence of diffusion algorithm although some links among normal neighbors may be cut \cite{Sayed2013Diffusion,Nassif2020Multitask}.

The proposed resilient DLMG (RDLMG) algorithm is summarized in Table \ref{tab:RDLMG}. As noted above, the proposed RDLMG algorithm possesses the robustness under impulsive interferences and the resilience against Byzantine attacks, which ensures that all normal nodes achieve their ideal states. In the following, we will present the theoretical performance analyses of RDLMG in the mean and mean-square senses.
\begin{table}[!ht]
\centering
\renewcommand{\arraystretch}{1.3}
\setlength{\abovecaptionskip}{2pt}
\caption{Summary of the RDLMG algorithm.}
\label{tab:RDLMG}
\resizebox{\linewidth}{!}{
\begin{tabular}{l}
\hline
\textbf{Initialization:} $\bm{w}_i(0), \gamma_{j,i}^2(0), \mu_i, \nu_i, \lambda$\\
\hline
  1:\hspace{6pt}\textbf{for} $n = 0,1,2,\dots$ \textbf{do}\\
  2:\hspace{12pt}\textbf{for} $i = 1,2,\dots$ \textbf{do}\\
  3:\hspace{18pt}$e_i(n) = d_i(n) - \bm{u}_i^\mathrm{T}(n) \bm{w}_i(n)$\\
  4:\hspace{18pt}$\bm{\psi}_i(n+1) = \bm{w}_i(n) + \mu_i \frac{1}{[1 + \lambda e_i^2(n)]^2} e_i(n) \bm{u}_i(n)$\\
  5:\hspace{12pt}\textbf{end for} \\
  6:\hspace{12pt}\textbf{for} $i = 1,2,\dots$ \textbf{do}\\
  7:\hspace{18pt}$\gamma_{j,i}^2(n+1) = (1-\nu_i)\gamma_{j,i}^2(n) + \nu_i\|\bm{\psi}_j(n+1) - \bm{w}_i(n)\|_2^2,j \in \mathcal{N}_i$\\
  8:\hspace{18pt}$Q_i(\bm{\psi}_j(n+1)) = \mathbb{E} \{  [d_i(n+1) - \bm{u}_i^\mathrm{T}(n+1) \bm{\psi}_j(n+1)]^2 \},j \in \mathcal{N}_i$\\
  9:\hspace{18pt}$c_{j,i}(n+1) =  \gamma_{j,i}^{-4}(n+1)Q_i(\bm{\psi}_j(n+1)),j \in \mathcal{N}_i$\\
  10:\hspace{14pt}$\mathcal{R}_i(n+1) = \{j \mid \arg \max_{j \in \mathcal{N}_i} c_{j,i}(n+1) \}$\\
  11:\hspace{14pt}$a_{j,i}(n+1) =
                  \left\{\begin{aligned}
                  &\frac{\gamma_{j,i}^{-2}(n+1)}{\sum_{ \scriptscriptstyle \ell \in \mathcal{N}_i \backslash \mathcal{R}_i(n+1)} \gamma_{\ell,i}^{-2}(n+1)},  & &  j \in \mathcal{N}_i \backslash \mathcal{R}_i(n+1)\\
                  &0, & &  \text{otherwise}
                  \end{aligned}\right. $ \\
  12:\hspace{14pt}$\bm{w}_i(n+1) = \sum_{\scriptscriptstyle j \in \mathcal{N}_i \backslash \mathcal{R}_i(n+1)} a_{j,i}(n+1) \bm{\psi}_j(n+1)$\\
  13:\hspace{8pt}\textbf{end for} \\
  14:\hspace{2pt}\textbf{end for} \\
\hline
\end{tabular}  }
\end{table}

\emph{Remark 3:} In fact, if the multi-task network is an $F$-local network \cite{LeBlanc2013Resilient}, then the RDLMG algorithm is also resilient to any number $F$ of Byzantine adversaries. From the perspective of resilience, we can assume that there can be at most $F$ Byzantine nodes among the neighbors of a normal node. In resilient multi-task diffusion, the $F$ extreme values will then be removed from each node, and the selection of $F$ does not affect the convergence, but it influences the steady-state performance.

In Table \ref{tab:complexity}, we summarize the computational complexity of the DLMS, DLMG, RDLMS and RDLMG algorithms for node $i$ per time instant $n$, where $n_i$ denotes the cardinality of $\mathcal{N}_i$. It can be seen that the resilient algorithms with the MSR mechanism have a significant increase of complexity as compared to DLMS and DLMG. In comparison with RDLMS, the proposed RDLMG algorithm slightly increases complexity caused by using the GM estimator, however it can be justified by the improvement of performance.

\begin{table}[!ht]
\centering
\renewcommand{\arraystretch}{1.3}
\setlength{\abovecaptionskip}{2pt}
\caption{Computational complexity of distributed algorithms for node $i$ per time instant.}
\label{tab:complexity}
\resizebox{\linewidth}{!}{
\begin{tabular}{cccc}
\hline
\textbf{Algorithms} & \textbf{Multiplications} & \textbf{Additions} & \textbf{Other operations} \\
\hline
  DLMS   & $n_iM+2M+1$     & $n_iM+M$              &--                    \\
  DLMG   & $n_iM+2M+5$     & $n_iM+M+1$            &--                   \\
  RDLMS  & $n_i(3M+7)+M+1$ & $2n_i(M+1)+2M-2$      & \makecell{1 expectation \\1 argmax operation}   \\
  RDLMG  & $n_i(3M+7)+M+5$ & $2n_i(M+1)+2M-1$      & \makecell{1 expectation \\1 argmax operation}   \\
\hline
\end{tabular} }
\end{table}

\section{Stability Analysis}

In this section, we present a statistical analysis of the proposed RDLMG algorithm in terms of mean performance, which establishes conditions for its stability. To proceed with the mathematical analysis, the following assumptions are commonly made.

\emph{Assumption 2:} $\eta_i(n)$ is a zero-mean i.i.d. noise of variance $\sigma_{\eta_i}^2$. It is customary to assume that $\eta_i(n)$ is modeled as a contaminated Gaussian (CG) noise for analyzing impulsive interferences \cite{Chen2018Diffusion,Zhou2011New}, whose model is described as
\begin{align}
\eta_i(n) = v_i(n) + b_i(n)g_i(n)
\end{align}
where $v_i(n)$ and $g_i(n)$ are both zero-mean Gaussian sequences with variances $\sigma_{v_i}^2$ and $\sigma_{g_i}^2$, respectively, and $\sigma_{g_i}^2 \gg \sigma_{v_i}^2$. $b_i(n)$ is a Bernoulli sequence whose value is either zero or one with probabilities $\mathrm{Pr}\{b_i(n)=1\}=p_i$ and $\mathrm{Pr}\{b_i(n)=0\}=1-p_i$. Accordingly, we note that the additive noise $\eta_i(n)$ is modeled as a CG noise with zero-mean and variance $\sigma_{\eta_i}^2 = \sigma_{v_i}^2 + p_i \sigma_{g_i}^2$, where $p_i$ represents the occurrence probability of impulsive interferences.

\emph{Assumption 3:} The regression vectors $\bm u_i(n)$ arise from a zero-mean random process that is temporally stationary, white, and independent over space with $\mathbb{E}\{\bm u_i(n)\bm u_i^\mathrm{T}(n)\}>0$. This is the well-known independence assumption \cite{Sayed2014Adaptation,Chen2015Diffusion}.

\emph{Assumption 4:} The scale function $f_i(e_i(n))$ is statistically independent of other random variables. This error nonlinearity can be regarded as a variable step size term to feasibly enable tractable analysis \cite{Ma2016Diffusion,Lee2015Variable}.

\emph{Assumption 5:} The adjacency matrix $\bm A(n)$ is statistically independent of $\bm u_i(n)$ and $\bm w_i(n)$. It is customary in the context of general adaptive combiners \cite{Nassif2016Multitask,Takahashi2010Diffusion}, which allows to simplify the analysis.

Subtracting both sides of \eqref{eq:psi} from $\bm{w}_i^\mathrm{o}$ yields
\begin{align}
\tilde{\bm{\psi}}_i(n+1) &=\tilde{\bm{w}}_i(n) - \mu_i \frac{1}{[1 + \lambda e_i^2(n)]^2} e_i(n) \bm{u}_i(n)   \label{eq:psi_error}\\
\tilde{\bm{w}}_i(n+1) &= \sum_{ \scriptscriptstyle j \in \mathcal{N}_i\setminus \mathcal{R}_i(n+1) } a_{j,i}(n+1) \tilde{\bm{\psi}}_j(n+1)  \label{eq:w_error}
\end{align}
where $\tilde{\bm{w}}_i(n) = \bm{w}_i^\mathrm{o} - \bm{w}_i(n)$ and $\tilde{\bm{\psi}}_i(n) = \bm{w}_i^\mathrm{o} - \bm{\psi}_i(n)$ are the estimation error and the intermediate estimation error, respectively. Substituting \eqref{eq:d} into \eqref{eq:e}, we can get $e_i(n) = \bm{u}_i^\mathrm{T}(n) \tilde{\bm{w}}_i(n) + \eta_i(n)$. Then, \eqref{eq:psi_error} can be rewritten as
\begin{align}
&~\tilde{\bm{\psi}}_i(n+1) \nonumber \\
=&~\tilde{\bm{w}}_i(n) - \mu_i f_i(e_i(n)) \bm{u}_i(n) [\bm{u}_i^\mathrm{T}(n) \tilde{\bm{w}}_i(n) + \eta_i(n)]  \nonumber \\
=&~[\bm I_M - \mu_i f_i(e_i(n)) \bm U_i(n) ]\tilde{\bm{w}}_i(n) - \mu_i f_i(e_i(n)) \bm h_i(n)
\end{align}
where $\bm I_M$ is an identity matrix of size $M \times M$, and $\bm U_i(n) = \bm{u}_i(n)\bm{u}_i^\mathrm{T}(n) \in \mathbb{R}^{M \times M}$ and $\bm h_i(n) = \bm{u}_i(n)\eta_i(n) \in \mathbb{R}^M$.

In fact, due to the presence of Byzantine nodes, the number of all normal nodes is $|\mathcal{V}\backslash\mathcal{V}^{\mathrm{a}}|$ over the graph $\mathcal{G}$, where $|\cdot|$ denotes the cardinality of its set argument. We use the number $N$ to denote $|\mathcal{V}\backslash\mathcal{V}^{\mathrm{a}}|$ for convenience. Some global quantities across all normal nodes can be put in the collective forms
\begin{align}
\tilde{\bm{w}}(n)&= \mathrm{col}\{\tilde{\bm{w}}_1(n),\tilde{\bm{w}}_2(n),\dots,\tilde{\bm{w}}_N(n)  \}\nonumber\\
\tilde{\bm{\psi}}(n)&= \mathrm{col}\{\tilde{\bm{\psi}}_1(n),\tilde{\bm{\psi}}_2(n),\dots,\tilde{\bm{\psi}}_N(n)  \}\nonumber\\
\bm h(n) & = \mathrm{col}\{\bm h_1(n),\bm h_2(n),\dots,\bm h_N(n)   \}\nonumber\\
\bm U(n) & = \mathrm{diag}\{\bm U_1(n),\bm U_2(n),\dots,\bm U_N(n)   \}\nonumber\\
\bm {\mathcal{M}} & = \mathrm{diag}\{\mu_1,\mu_2,\dots,\mu_N \}\otimes \bm I_M \nonumber\\
\bm {\mathcal{F}}(n) & = \mathrm{diag}\{f_1(e_1(n)),f_2(e_2(n)),\dots,f_N(e_N(n)) \}\otimes \bm I_M \nonumber\\
\bm {\mathcal{A}}(n) & = \bm A(n)\otimes \bm I_M
\end{align}
where each column vector $\mathrm{col}\{\cdot\}$ with $N \times 1$ block vectors is of size $NM \times 1$, and each diagonal matrix $\mathrm{diag}\{\cdot\}$ with $N \times N$ block diagonal matrices is of size $NM \times NM$, and $\otimes$ is the Kronecker product, respectively. Besides, $\bm A (n) \in \mathbb{R}^{N \times N}$ is also the adjacency matrix across all normal nodes and $\bm A^\mathrm{T}(n) \bm 1_N = \bm 1_N$ still holds.

We can describe the collective relations for \eqref{eq:psi_error} and \eqref{eq:w_error} in the following forms
\begin{align}
\tilde{\bm{\psi}}(n+1) &= [\bm I_{NM} - \bm {\mathcal{M}} \bm {\mathcal{F}}(n) \bm U(n)]\tilde{\bm{w}}(n)- \bm {\mathcal{M}} \bm {\mathcal{F}}(n) \bm h(n) \nonumber\\
\tilde{\bm{w}}(n+1) & = \bm {\mathcal{A}}^\mathrm{T}(n+1) \tilde{\bm{\psi}}(n+1)   .
\end{align}
Then, the networked estimation error for all normal nodes can be obtained as
\begin{align}
\tilde{\bm{w}}(n+1)  =&~\bm {\mathcal{A}}^\mathrm{T}(n+1) [\bm I_{NM} - \bm {\mathcal{M}} \bm {\mathcal{F}}(n) \bm U(n)]\tilde{\bm{w}}(n) \nonumber \\
&- \bm {\mathcal{A}}^\mathrm{T}(n+1) \bm {\mathcal{M}} \bm {\mathcal{F}}(n) \bm h(n)   .  \label{eq:network_w_error}
\end{align}

Considering the mathematical expectation on both sides of \eqref{eq:network_w_error} under Assumptions 2 and 3, it yields
\begin{align}
&~\mathbb{E}\{\tilde{\bm{w}}(n+1)\}  \nonumber\\
=&~\mathbb{E}\{\bm {\mathcal{A}}^\mathrm{T}(n+1) [\bm I_{NM} - \bm {\mathcal{M}} \bm {\mathcal{F}}(n) \bm U(n)]\tilde{\bm{w}}(n) \} \nonumber\\
&- \mathbb{E}\{\bm {\mathcal{A}}^\mathrm{T}(n+1) \bm {\mathcal{M}} \bm {\mathcal{F}}(n) \bm r(n)\}  \nonumber\\
= &~\mathbb{E}\{\bm {\mathcal{A}}^\mathrm{T}(n+1) [\bm I_{NM} - \bm {\mathcal{M}} \bm {\mathcal{F}}(n) \bm U(n)] \} \mathbb{E}\{\tilde{\bm{w}}(n)\}  .
\end{align}
Using the instantaneous approximation $\mathbb{E}\{\bm {\mathcal{A}}^\mathrm{T}(n+1)\} \approx \bm {\mathcal{A}}^\mathrm{T}(n+1)$, as well as Assumptions 4 and 5, we have
\begin{align}
&~\mathbb{E}\{\bm {\mathcal{A}}^\mathrm{T}(n+1) [\bm I_{NM} - \bm {\mathcal{M}} \bm {\mathcal{F}}(n) \bm U(n)] \} \mathbb{E}\{\tilde{\bm{w}}(n)\} \nonumber\\
\approx &~ \bm {\mathcal{A}}^\mathrm{T}(n+1)[\bm I_{NM} - \bm {\mathcal{M}} \mathbb{E}\{ \bm {\mathcal{F}}(n)\} \mathbb{E}\{\bm U(n)\}] \mathbb{E}\{\tilde{\bm{w}}(n)\}  .
\end{align}
Then, we can further obtain
\begin{align}\label{eq:Exp_network_w_error}
\mathbb{E}\{\tilde{\bm{w}}(n+1)\} = \bm \Gamma_n \mathbb{E}\{\tilde{\bm{w}}(n)\}
\end{align}
where we denote
\begin{align}
\bm \Gamma_n \triangleq \bm {\mathcal{A}}^\mathrm{T}(n+1)[\bm I_{NM} - \bm {\mathcal{M}} \mathbb{E}\{ \bm {\mathcal{F}}(n)\} \mathbb{E}\{\bm U(n)\}]  .
\end{align}
From \eqref{eq:Exp_network_w_error}, we have the following result of mean stability for the RDLMG algorithm.

\emph{Theorem 1:} The RDLMG algorithm converges in the mean for any initial condition, if the step size $\mu_i$ is chosen to satisfy
\begin{align}\label{eq:mean_mu}
0< \mu_i <\frac{2}{\mathbb{E}\{f_i(e_i(n))\} \lambda_\mathrm{max}( \mathbb{E}\{\bm U_i(n)\})}
\end{align}
where $\lambda_\mathrm{max}(\cdot)$ denotes the maximum eigenvalue of its matrix argument.

\emph{Proof:} Repeatedly iterating \eqref{eq:Exp_network_w_error} leads to
\begin{align}\label{eq:Ite_network_w_error}
\mathbb{E}\{\tilde{\bm{w}}(n+1)\} = \prod_{\ell =0}^{n} \bm \Gamma_\ell \mathbb{E}\{\tilde{\bm{w}}(0)\}   .
\end{align}

Let the block maximum norm of a $N \times 1$ block column vector $\bm x = \mathrm{col}\{\bm x_1,\bm x_2,\dots,\bm x_N\} \in \mathbb{R}^{NM}$ with block entries of size $M \times 1$ each \cite{Takahashi2010Diffusion}, define as
\begin{align}
\|\bm x\|_{\mathrm{b},\infty} \triangleq \max_{1\le i \le N}\|\bm x_i \|_2  .
\end{align}
Accordingly, let the block maximum norm of a $N \times N$ block matrix $\bm {\mathcal{X}} \in \mathbb{R}^{NM \times NM}$ with block entries of size $M \times M$ each, define as
\begin{align}
\|\bm {\mathcal{X}} \|_{\mathrm{b},\infty} \triangleq \max_{\bm x \neq 0} \frac{ \|\bm {\mathcal{X}} \bm x_i \|_{\mathrm{b},\infty} }{ \|\bm x_i \|_{\mathrm{b},\infty} }   .
\end{align}
Obviously, $\| \bm {\mathcal{A}}^\mathrm{T}(n)\|_{\mathrm{b},\infty} =1$ holds. Calculating the block maximum norm on both sides of \eqref{eq:Ite_network_w_error} yields
\begin{align}
\| \mathbb{E}\{\tilde{\bm{w}}(n+1)\} \|_{\mathrm{b},\infty} &= \bigg\| \prod_{\ell =0}^{n} \bm \Gamma_\ell \mathbb{E}\{\tilde{\bm{w}}(0)\} \bigg\|_{\mathrm{b},\infty}  \nonumber\\
&\leq \prod_{\ell =0}^{n} \| \bm \Gamma_\ell \|_{\mathrm{b},\infty} \cdot \| \mathbb{E}\{\tilde{\bm{w}}(0)\} \|_{\mathrm{b},\infty} \nonumber\\
&\leq \Big (\max_{0 \le \ell \le n} \| \bm \Gamma_\ell \|_{\mathrm{b},\infty}\Big)^n \cdot \| \mathbb{E}\{\tilde{\bm{w}}(0)\} \|_{\mathrm{b},\infty}  .
\end{align}
It implies that $\| \mathbb{E}\{\tilde{\bm{w}}(n+1)\} \|_{\mathrm{b},\infty}$ will converge if $\| \bm \Gamma_n \|_{\mathrm{b},\infty}<1$. We thus consider
\begin{align}
&~\|\bm \Gamma_n \|_{\mathrm{b},\infty}\nonumber\\
=&~\| \bm {\mathcal{A}}^\mathrm{T}(n+1)[\bm I_{NM} - \bm {\mathcal{M}} \mathbb{E}\{ \bm {\mathcal{F}}(n)\} \mathbb{E}\{\bm U(n)\}] \|_{\mathrm{b},\infty} \nonumber\\
\leq&~\| \bm {\mathcal{A}}^\mathrm{T}(n+1)\|_{\mathrm{b},\infty} \cdot \|[\bm I_{NM} - \bm {\mathcal{M}} \mathbb{E}\{ \bm {\mathcal{F}}(n)\} \mathbb{E}\{\bm U(n)\}] \|_{\mathrm{b},\infty} \nonumber\\
=&~\|[\bm I_{NM} - \bm {\mathcal{M}} \mathbb{E}\{ \bm {\mathcal{F}}(n)\} \mathbb{E}\{\bm U(n)\}] \|_{\mathrm{b},\infty} \nonumber\\
\leq&~\max_{1 \le i \le N} \vertiii \bm I_M - \mu_i \mathbb{E}\{f_i(e_i(n))\} \mathbb{E}\{\bm U_i(n)\}  \vertiii_2 \nonumber\\
<&~1
\end{align}
where $\vertiii \cdot \vertiii_2$ denotes the induced $l_2$-norm (or spectral norm) of its matrix argument.

A sufficient condition for $\| \bm \Gamma_n \|_{\mathrm{b},\infty}<1$ is to ensure $|1-\mu_i \mathbb{E}\{f_i(e_i(n))\} \lambda_\mathrm{max}( \mathbb{E}\{\bm U_i(n)\})|<1$. This leads us to the condition in \eqref{eq:mean_mu}. Furthermore, if the $l_1$-norm of the estimate $\bm w_i(n)$ is bounded by $\beta_i$, then it holds that \cite{Chen2015Convergence}
\begin{align}
|e_i(n)| &= |d_i(n) - \bm{u}_i^\mathrm{T}(n) \bm{w}_i(n)| \nonumber\\
&\le \|\bm{w}_i(n)\|_1 \cdot \|\bm{u}_i(n)\|_1 + |d_i(n)| \nonumber\\
&\le \beta_i \|\bm{u}_i(n)\|_1 + |d_i(n)|
\end{align}
where $\|\cdot\|_1$ denotes the $l_1$-norm (or sum norm) of its vector argument. This leads to the sufficient condition as
\begin{align}\label{eq:mean_mu2}
0< \mu_i <\frac{2}{\mathbb{E}\{f_i(\beta_i \|\bm{u}_i(n)\|_1 + |d_i(n)|)\} \lambda_\mathrm{max}( \mathbb{E}\{\bm U_i(n)\})}   .
\end{align}
Also, when $n \to \infty$, it holds that $\| \mathbb{E}\{\tilde{\bm{w}}(n+1)\} \|_{\mathrm{b},\infty} \to 0$, i.e., $\| \mathbb{E}\{\tilde{\bm{w}}(n+1)\} \|_2 \to 0$ for $n \to \infty$,which further implies $\lim_{n \to \infty}\mathbb{E}\{\bm{w}_i(n+1)\} = \bm{w}_i^\mathrm{o}$. As a consequence, the proposed distributed strategy can ensure mean stability if the step size $\mu_i$ is selected to satisfy the above condition. This completes the proof.  \hfill $\blacksquare$

\section{Mean-Square Performance Analysis}

We present the mean-square performance analysis of the RDLMG algorithm in this section. Let the covariance matrix of the networked estimation error $\tilde{\bm{w}}(n)$ for all normal nodes be denoted as
\begin{align}
\bm {\mathscr{W}}_{n} \triangleq \mathbb{E}\{ \tilde{\bm{w}}(n)\tilde{\bm{w}}^{\mathrm{T}}(n)  \} \in \mathbb{R}^{NM \times NM}
\end{align}
with $N \times N$ block diagonal matrix, whose individual block is $\bm {\mathscr{W}}_i(n) \triangleq \mathbb{E}\{\tilde{\bm{w}}_i(n)\tilde{\bm{w}}_i^{\mathrm{T}}(n)\} \in \mathbb{R}^{M \times M}$ at each node $i$.

Post-multiplying \eqref{eq:network_w_error} by its transpose and taking the expectation operation for the relation under Assumptions 2-5, we can deduce
\begin{align}
&~\bm {\mathscr{W}}_{n+1} \nonumber\\
=&~ \mathbb{E}\{ \tilde{\bm{w}}(n+1)\tilde{\bm{w}}^{\mathrm{T}}(n+1)  \} \nonumber\\
=&~ \mathbb{E} \big \{ \bm {\mathcal{A}}^\mathrm{T}(n+1) [\bm I_{NM} - \bm {\mathcal{M}} \bm {\mathcal{F}}(n) \bm U(n)]\tilde{\bm{w}}(n)  \nonumber\\
&~\times \tilde{\bm{w}}^{\mathrm{T}}(n) [\bm I_{NM} - \bm {\mathcal{M}} \bm {\mathcal{F}}(n) \bm U(n)]^{\mathrm{T}}\bm {\mathcal{A}} (n+1) \nonumber\\
&~+ \bm {\mathcal{A}}^\mathrm{T}(n+1) \bm {\mathcal{M}} \bm {\mathcal{F}}(n) \bm h(n) \bm h^{\mathrm{T}}(n) \bm {\mathcal{F}}(n)\bm {\mathcal{M}} \bm {\mathcal{A}} (n+1)\big\} \nonumber\\
=&~\bm {\mathscr{A}}_{n+1}^\mathrm{T} [\bm I_{NM} - \bm {\mathcal{M}} \bm {\mathscr{F}}_n \bm {\mathscr{U}}] \bm {\mathscr{W}}_n [\bm I_{NM} - \bm {\mathcal{M}} \bm {\mathscr{F}}_n \bm {\mathscr{U}}]^\mathrm{T} \bm {\mathscr{A}}_{n+1} \nonumber\\
&~+\bm {\mathscr{A}}_{n+1}^\mathrm{T} \bm {\mathcal{M}} \bm {\mathscr{F}}_n \bm {\mathscr{H}} \bm {\mathscr{F}}_n \bm {\mathcal{M}} \bm {\mathscr{A}}_{n+1} \label{eq:vec_w}
\end{align}
where we denote
\begin{align}
\bm {\mathscr{A}}_{n+1} \triangleq&~ \mathbb{E}\{ \bm {\mathcal{A}} (n+1) \} \nonumber\\
\bm {\mathscr{F}}_n \triangleq&~ \mathbb{E}\{ \bm {\mathcal{F}} (n) \} \nonumber\\
\bm {\mathscr{U}} \triangleq&~ \mathbb{E}\{ \bm U (n) \} \nonumber\\
\bm {\mathscr{H}} \triangleq&~ \mathbb{E}\{ \bm h (n)\bm h^{\mathrm{T}} (n) \}\nonumber\\
=&~\mathrm{diag}\{ \sigma_{\eta_1}^2 \mathbb{E}\{ \bm U_1 (n) \}, \dots, \sigma_{\eta_N}^2 \mathbb{E}\{ \bm U_N (n) \} \}   .
\end{align}

The matrix property $\mathrm{vec}(\bm {\mathscr{X}} \bm {\mathscr{Y}} \bm {\mathscr{Z}}) = (\bm {\mathscr{Z}}^{\mathrm{T}} \otimes  \bm {\mathscr{X}})\mathrm{vec}( \bm {\mathscr{Y}} )$ remains for $\forall \bm {\mathscr{X}},\bm {\mathscr{Y}},\bm {\mathscr{Z}}$ with compatible dimensions \cite{Sayed2014Adaptation}, where the vectorization operation $\mathrm{vec}(\cdot)$ represents a column vector obtained by stacking the columns of its matrix argument. Taking the vectorization operation on both sides of \eqref{eq:vec_w}, we can get
\begin{align}
&~\mathrm{vec}( \bm {\mathscr{W}}_{n+1} ) \nonumber\\
=&~ \big[\big ([\bm I_{NM} - \bm {\mathcal{M}} \bm {\mathscr{F}}_n \bm {\mathscr{U}}]^\mathrm{T} \bm {\mathscr{A}}_{n+1} \big)^\mathrm{T} \nonumber\\
&~\otimes \big( \bm {\mathscr{A}}_{n+1}^\mathrm{T} [\bm I_{NM} - \bm {\mathcal{M}} \bm {\mathscr{F}}_n \bm {\mathscr{U}}] \big)\big]\mathrm{vec}( \bm {\mathscr{W}}_{n} )\nonumber\\
&~+ \big[\big( \bm {\mathscr{F}}_n \bm {\mathcal{M}} \bm {\mathscr{A}}_{n+1}\big)^{\mathrm{T}}\otimes \big( \bm {\mathscr{A}}_{n+1}^\mathrm{T} \bm {\mathcal{M}} \bm {\mathscr{F}}_n \big)\big]\mathrm{vec}( \bm {\mathscr{H}} ) \nonumber\\
=&~\big[\big( \bm {\mathscr{A}}_{n+1}^\mathrm{T} [\bm I_{NM} - \bm {\mathcal{M}} \bm {\mathscr{F}}_n \bm {\mathscr{U}}] \big) \otimes \big( \bm {\mathscr{A}}_{n+1}^\mathrm{T} [\bm I_{NM} - \bm {\mathcal{M}} \bm {\mathscr{F}}_n \bm {\mathscr{U}}] \big)\big]\nonumber\\
&~\times \mathrm{vec}( \bm {\mathscr{W}}_{n} )+ \big[\big( \bm {\mathscr{A}}_{n+1}^\mathrm{T} \bm {\mathcal{M}} \bm {\mathscr{F}}_n \big)\otimes\big( \bm {\mathscr{A}}_{n+1}^\mathrm{T} \bm {\mathcal{M}} \bm {\mathscr{F}}_n \big)\big]\mathrm{vec}( \bm {\mathscr{H}} )  .
\end{align}
We can then arrange
\begin{align}
&~\bm \Phi_n \nonumber\\
\triangleq &~ \big[ \bm {\mathscr{A}}_{n+1}^\mathrm{T} (\bm I_{NM} - \bm {\mathcal{M}} \bm {\mathscr{F}}_n \bm {\mathscr{U}}) \big] \otimes \big[ \bm {\mathscr{A}}_{n+1}^\mathrm{T} (\bm I_{NM} - \bm {\mathcal{M}} \bm {\mathscr{F}}_n \bm {\mathscr{U}}) \big] \nonumber\\
=&~(\bm {\mathscr{A}}_{n+1}^\mathrm{T} \otimes \bm {\mathscr{A}}_{n+1}^\mathrm{T}) \big[(\bm I_{NM} - \bm {\mathcal{M}} \bm {\mathscr{F}}_n \bm {\mathscr{U}}) \otimes (\bm I_{NM} - \bm {\mathcal{M}} \bm {\mathscr{F}}_n \bm {\mathscr{U}}) \big]\nonumber\\
=&~(\bm {\mathscr{A}}_{n+1}^\mathrm{T} \otimes \bm {\mathscr{A}}_{n+1}^\mathrm{T})\big[ \bm I_{N^2M^2}  - \bm I_{NM} \otimes (\bm {\mathcal{M}} \bm {\mathscr{F}}_n \bm {\mathscr{U}})\nonumber\\
&- (\bm {\mathcal{M}} \bm {\mathscr{F}}_n \bm {\mathscr{U}}) \otimes \bm I_{NM} + [(\bm {\mathcal{M}} \bm {\mathscr{F}}_n) \otimes(\bm {\mathcal{M}} \bm {\mathscr{F}}_n)](\bm {\mathscr{U}}\otimes \bm {\mathscr{U}})\big]   .
\end{align}
Then, it can be established that
\begin{align}
\mathrm{vec}( \bm {\mathscr{W}}_{n+1} )=&~\bm \Phi_n \mathrm{vec}( \bm {\mathscr{W}}_n )+ (\bm {\mathscr{A}}_{n+1}^\mathrm{T} \otimes \bm {\mathscr{A}}_{n+1}^\mathrm{T})  \nonumber\\
&~\times [ (\bm {\mathcal{M}} \bm {\mathscr{F}}_n) \otimes (\bm {\mathcal{M}} \bm {\mathscr{F}}_n) ]\mathrm{vec}( \bm {\mathscr{H}} )   . \label{eq:vec_w_error}
\end{align}

As a performance metric in the mean-square sense, the MSD at the normal node $i \in \mathcal{V}\backslash\mathcal{V}^{\mathrm{a}}$ is defined as
\begin{align}
\mathrm{MSD}_i(n) = \mathrm{Tr}(\bm {\mathscr{W}}_i(n)) = \mathrm{Tr}(\mathbb{E}\{\tilde{\bm{w}}_i(n)\tilde{\bm{w}}_i^{\mathrm{T}}(n)\})
\end{align}
where $\mathrm{Tr}(\cdot)$ represents the trace of a matrix. Consider that the clustered multi-task network is resilient, but it may be decomposed into $\kappa$ connected sub-networks by executing the RDLMG algorithm. Each set of normal nodes of final sub-networks is denoted as $\mathcal{S}_\ell, \ell \in\{1,2,\dots,\kappa\}$ with satisfying $\bigcup_{\ell=1}^\kappa \mathcal{S}_\ell = \mathcal{V} \backslash \mathcal{V}^\mathrm{a}$. The MSD for each sub-network $\mathcal{S}_\ell$ is
\begin{align}
\mathrm{MSD}_{\mathcal{S}_\ell}(n) = \frac{1}{|\mathcal{S}_\ell|}\sum_{i \in \mathcal{S}_\ell} \mathrm{MSD}_i(n)  .
\end{align}
Accordingly, the networked MSD for all normal nodes over the clustered multi-task network is defined as
\begin{align}
\mathrm{MSD}(n) &= \frac{1}{| \mathcal{V} \backslash \mathcal{V}^\mathrm{a} |} \sum_{\ell=1}^\kappa \mathrm{MSD}_{\mathcal{S}_\ell}(n) \cdot |\mathcal{S}_\ell| \nonumber\\
&= \frac{1}{| \mathcal{V} \backslash \mathcal{V}^\mathrm{a} |} \mathrm{Tr}(\bm {\mathscr{W}}_{n}).
\end{align}

Taking into account the steady-state performance for $n \to \infty$, we can deduce from \eqref{eq:vec_w_error} as
\begin{align}
\mathrm{vec}( \bm {\mathscr{W}}_\infty )=&~ ( \bm I_{N^2M^2} -  \bm \Phi_\infty )^{-1} (\bm {\mathscr{A}}_\infty^\mathrm{T} \otimes \bm {\mathscr{A}}_\infty^\mathrm{T}) \nonumber\\
&~\times[ (\bm {\mathcal{M}} \bm {\mathscr{F}}_\infty) \otimes (\bm {\mathcal{M}} \bm {\mathscr{F}}_\infty) ]\mathrm{vec}( \bm {\mathscr{H}} ) \label{eq:vec_w_inf}
\end{align}
with
\begin{align}
&~\bm \Phi_\infty \nonumber\\
= &~(\bm {\mathscr{A}}_\infty^\mathrm{T} \otimes \bm {\mathscr{A}}_\infty^\mathrm{T})\big[ \bm I_{N^2M^2}  - \bm I_{NM} \otimes (\bm {\mathcal{M}} \bm {\mathscr{F}}_\infty \bm {\mathscr{U}})\nonumber\\
&- (\bm {\mathcal{M}} \bm {\mathscr{F}}_\infty \bm {\mathscr{U}}) \otimes \bm I_{NM} + [(\bm {\mathcal{M}} \bm {\mathscr{F}}_\infty) \otimes(\bm {\mathcal{M}} \bm {\mathscr{F}}_\infty)](\bm {\mathscr{U}}\otimes \bm {\mathscr{U}})\big]  .
\end{align}

In the following, we will calculate the values of $\bm {\mathscr{A}}_\infty$ and $\bm {\mathscr{F}}_\infty$. Considering the steady-state for $n \to \infty$, we have $e_i(n) = \bm{u}_i^\mathrm{T}(n) \tilde{\bm{w}}_i(n) + \eta_i(n) \approx \eta_i(n)$.
According to \eqref{eq:psi}, we can get
\begin{align}\label{eq:psi_w}
\bm{\psi}_i(n+1) = \bm{w}_i^\mathrm{o} + \mu_i \frac{1}{[1 + \lambda e_i^2(n)]^2} e_i(n) \bm{u}_i(n)  .
\end{align}
Subtracting $\bm{w}_i^\mathrm{o}$ from both sides of \eqref{eq:psi_w}, and taking expectation for the energies on both sides of the relation, as well as Assumption 2, it yields
\begin{align}
&~\mathbb{E} \{ \|\bm{\psi}_i(n+1) - \bm{w}_i^\mathrm{o}\|_2^2\} \nonumber\\
=&~\mu_i^2 \mathrm{Tr}(\mathbb{E}\{\bm u_i(n)\bm u_i^{\mathrm{T}}(n)\})\cdot\mathbb{E}\bigg \{\frac{e_i^2(n)}{[1 + \lambda e_i^2(n)]^4}\bigg\}  . \label{eq:Exp_psi_w}
\end{align}
Since the small fluctuation of $|e_i(n)|$ \cite{Lu2018Performance} and the small enough Jensen gap \cite{Briat2011Convergence} at steady-state, we can make the approximation
\begin{align}
\lim_{n \to \infty}\mathbb{E}\bigg \{\frac{e_i^2(n)}{[1 + \lambda e_i^2(n)]^4}\bigg\} \approx \frac{\sigma_{\eta_i}^2}{(1+\lambda\sigma_{\eta_i}^2)^4}  .
\end{align}
Therefore, we can rewrite \eqref{eq:Exp_psi_w} for $n \to \infty$ as
\begin{align}
\mathbb{E} \{ \|\bm{\psi}_i(n+1) - \bm{w}_i^\mathrm{o}\|_2^2\}\approx\mu_i^2 \mathrm{Tr}(\mathbb{E}\{\bm U_i(n)\})\frac{\sigma_{\eta_i}^2}{(1+\lambda\sigma_{\eta_i}^2)^4}  .
\end{align}

At steady-state for $n \to \infty$, each node and its normal neighbors performing the same objective belong to the same sub-network, and thus $\bm{w}_i^\mathrm{o} = \bm{w}_j^\mathrm{o}, j \in \mathcal{N}_i$. Taking expectation on both sides of \eqref{eq:update_weights} yields
\begin{align}
&~\mathbb{E}\{ \gamma_{j,i}^2(n+1)\} \nonumber\\
=&~ (1-\nu_i)\mathbb{E}\{\gamma_{j,i}^2(n)\}+ \nu_i\mathbb{E}\{\|\bm{\psi}_j(n+1) - \bm{w}_j^\mathrm{o}\|_2^2\} \nonumber\\
=&~ (1-\nu_i)\mathbb{E}\{\gamma_{j,i}^2(n)\} + \nu_i \mu_j^2 \mathrm{Tr}(\mathbb{E}\{\bm U_j(n)\})\frac{\sigma_{\eta_j}^2}{(1+\lambda\sigma_{\eta_j}^2)^4}  .
\end{align}
By repeatedly iterating, we get
\begin{align}
\mathbb{E}\{ \gamma_{j,i}^{-2}(\infty)\} =  \frac{(1+\lambda\sigma_{\eta_j}^2)^4}{\mu_j^2 \mathrm{Tr}(\mathbb{E}\{\bm U_j(n)\})\sigma_{\eta_j}^2}  .
\end{align}
We can further obtain
\begin{small}
\begin{align}
\displaystyle \mathbb{E}\{a_{j,i}(\infty)\} =
\left\{\begin{aligned}
&\frac{\mathbb{E}\{\gamma_{j,i}^{-2}(\infty)\}}{\sum_{\scriptscriptstyle \ell \in \mathcal{N}_i \backslash \mathcal{R}_i(n+1)} \mathbb{E}\{\gamma_{\ell,i}^{-2}(\infty)\}},  & &  j \in \mathcal{N}_i\backslash \mathcal{R}_i(\infty) \\
&0, & &  \text{otherwise}  .
\end{aligned}\right.
\end{align}
\end{small}Then, $\bm A(\infty)= [a_{i,j}(\infty)]$ can be obtained directly.

As previously mentioned, we can make the approximation
\begin{align}
\mathbb{E}\{f_i(e_i(\infty))\} = \lim_{n \to \infty}\mathbb{E}\bigg \{\frac{1}{[1 + \lambda e_i^2(n)]^2}\bigg\} \approx \frac{1}{(1+\lambda\sigma_{\eta_i}^2)^2}.
\end{align}

As a consequence, the values of $\bm {\mathscr{A}}_\infty$ and $\bm {\mathscr{F}}_\infty$ can be computed as
\begin{align}\label{eq:A_inf}
\bm {\mathscr{A}}_\infty  &= \mathbb{E}\{\bm A(\infty)\otimes \bm I_M\} =[\mathbb{E}\{a_{i,j}(\infty)\}] \otimes \bm I_M
\end{align}
and
\begin{align}\label{eq:F_inf}
\bm {\mathscr{F}}_\infty &= \mathrm{diag}\{\mathbb{E}\{f_1(e_1(\infty))\},\dots,\mathbb{E}\{f_N(e_N(\infty))\} \}\otimes \bm I_M  .
\end{align}
Then, the theoretical value of $\mathrm{vec}( \bm {\mathscr{W}}_\infty )$ represented as \eqref{eq:vec_w_inf} can be provided by \eqref{eq:A_inf} and \eqref{eq:F_inf}.

Using the matrix property $\mathrm{Tr}(\bm {\mathscr{X}} \bm {\mathscr{Y}}) = \big[\mathrm{vec}(\bm {\mathscr{Y}}^{\mathrm{T}})\big]^{\mathrm{T}} \mathrm{vec}(\bm {\mathscr{X}})$ \cite{Sayed2014Adaptation}, the steady-state networked MSD for all normal nodes over the clustered multi-task network can be calculated as
\begin{align}
\mathrm{MSD}(\infty)=&~ \frac{1}{| \mathcal{V} \backslash \mathcal{V}^\mathrm{a} |} \mathrm{Tr}(\bm {\mathscr{W}}_\infty) \nonumber\\
=&~ \frac{1}{N} [\mathrm{vec}(\bm I_{NM})]^{\mathrm{T}} \mathrm{vec}(\bm {\mathscr{W}}_\infty) \nonumber\\
&~  \frac{1}{N} [\mathrm{vec}(\bm I_{NM})]^{\mathrm{T}} ( \bm I_{N^2M^2} -  \bm \Phi_\infty )^{-1} \nonumber\\
&~\times(\bm {\mathscr{A}}_\infty^\mathrm{T} \otimes \bm {\mathscr{A}}_\infty^\mathrm{T})[ (\bm {\mathcal{M}} \bm {\mathscr{F}}_\infty) \otimes (\bm {\mathcal{M}} \bm {\mathscr{F}}_\infty) ]\mathrm{vec}( \bm {\mathscr{H}} ) \label{eq:MSD_inf}
\end{align}
where the values of $\bm {\mathscr{A}}_\infty$ and $\bm {\mathscr{F}}_\infty$ are given by \eqref{eq:A_inf} and \eqref{eq:F_inf}.

\section{Numerical Results}

In this section, illustrative examples with applications of multi-target localization and multi-task spectrum sensing are performed to verify the proposed RDLMG algorithm. We also evaluate the LMS-based, NC-LMG, DLMG, and RDLMG algorithms, and compare their performances in the presence of impulsive interferences and Byzantine attacks.

\subsection{Multi-Target Localization}

We consider that the nodes are interested in locating multiple targets represented by the two-dimensional state vector $\bm w_i^\mathrm{o} = [x_i^\mathrm{o},y_i^\mathrm{o}]^\mathrm{T}$ in the Cartesian plane. The location of each node $i$ is denoted by $\bm p_i = [x_i,y_i]^\mathrm{T}$, and the distance between the node $i$ and its target $\bm w_i^\mathrm{o}$ is described by
\begin{align}
r_i^\mathrm{o} = \|\bm w_i^\mathrm{o} - \bm p_i\|_2 = \bm u_i^\mathrm{oT} (\bm w_i^\mathrm{o} - \bm p_i)
\end{align}
where $\bm u_i^\mathrm{o}=\frac{\bm w_i^\mathrm{o} - \bm p_i}{\|\bm w_i^\mathrm{o} - \bm p_i\|_2}$ represents the unit direction vector from node $i$ to its target \cite{Chen2014Multitask,Li2020Resilient}. In practice, the node has noisy observations of the distance and the direction vector towards the target, and thus the noisy measurements $\{r_i(n),\bm u_i(n)\}$ at time $n$ can be represented by
\begin{align}
r_i(n) = \bm u_i^\mathrm{T}(n) (\bm w_i^\mathrm{o} - \bm p_i) + \eta_i(n)
\end{align}
where $\eta_i(n)$ is the additive observation noise. Define the adjusted signal as $d_i(n) = r_i(n) + \bm u_i^\mathrm{T}(n)\bm p_i$. The linear model for the available measurement data $\{d_i(n),\bm u_i(n)\}$ is provided to estimate the location of the target $\bm w_i^\mathrm{o}$ as follows:
\begin{align}
d_i(n) = \bm{u}_i^\mathrm{T}(n) \bm{w}_i^\mathrm{o} + \eta_i(n)  .
\end{align}

\begin{figure}[!t]
\centering
\includegraphics[width=1.6in]{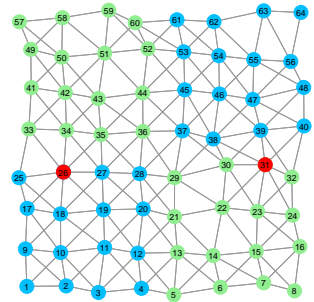}
\caption{Initial multi-task network topology.}
\label{fig:initial_network}
\vspace{-10pt}
\end{figure}

Consider a connected network with 64 nodes for the clustered multi-task scenario as shown in Fig. \ref{fig:initial_network}, which comprises of 62 normal nodes divided into two groups with blue and green nodes, and contains two Byzantine nodes with red nodes. Each normal node can be modeled as \eqref{eq:d} for target localization, and $\bm w_i(n) = [w_{i,1}(n),w_{i,2}(n)]^\mathrm{T}$ is the estimate of $\bm w_i^\mathrm{o}$. We set the multi-target localization of the two static targets as
\begin{align}
\bm w_i^\mathrm{o} =
\left\{\begin{aligned}
&[0.1,0.2]^\mathrm{T},  & &  \text{for node}~i~\text{depicted in blue}\\
&[0.7,0.8]^\mathrm{T},  & &  \text{for node}~i~\text{depicted in green}  .
\end{aligned}\right.
\end{align}

\begin{figure}[!t]
\centering
\includegraphics[width=2.5in]{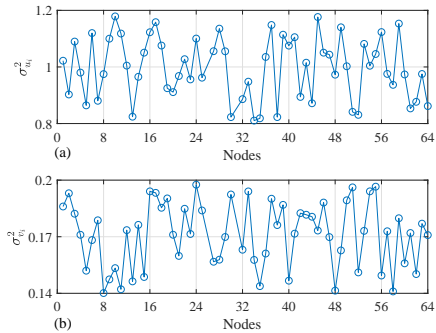}
\caption{Variances of input and noise signals for all normal nodes. (a) Values of $\sigma_{u_i}^2$. (b) Values of $\sigma_{v_i}^2$.}
\label{fig:nodes_values}
\vspace{-10pt}
\end{figure}

The two-dimensional regression vector at node $i$ is given by $\bm u_i(n)$, with $u_i(n)$ being extracted from the zero-mean Gaussian signal with variance $\sigma_{u_i}^2$. The additive noise $\eta_i(n)$ belongs to the zero-mean CG impulsive interference under Assumption 2, where $v_i(n)$ and $g_i(n)$ are both zero-mean Gaussian sequences with variances $\sigma_{v_i}^2$ and $\sigma_{g_i}^2 = 10^4\sigma_{v_i}^2$, and with the occurrence probability of impulsive interferences being set as $p_i = 0.01$. Fig. \ref{fig:nodes_values} illustrates the values of $\sigma_{u_i}^2$ and $\sigma_{v_i}^2$ for all normal nodes.

The Byzantine nodes aim to drive their neighboring normal nodes to estimate the location of the malicious target as $\bm w_i^\mathrm{a} = [0.4,0.5]^\mathrm{T}$, and the parameter is selected as $\mu_i^\mathrm{a}=0.01$ across the malicious nodes.

\begin{figure}[!t]
\centering
\includegraphics[width=1.6in]{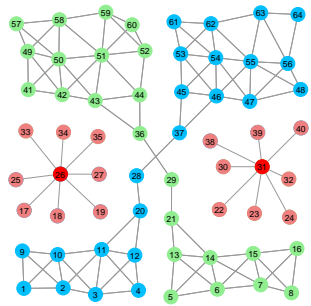}
\caption{Final network topology by executing DLMG.}
\label{fig:end_network}
\vspace{-10pt}
\end{figure}

\begin{figure}[!t]
\centering
\includegraphics[width=2.5in]{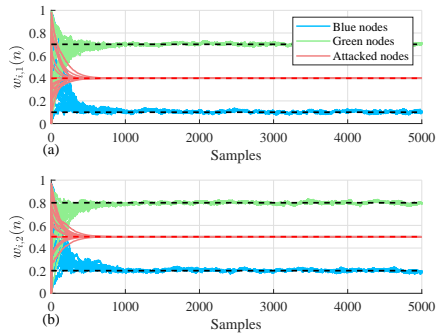}
\caption{Estimates of localization by executing DLMG. (a) $X$-coordinate. (b) $Y$-coordinate.}
\label{fig:w1w2_location_without}
\vspace{-10pt}
\end{figure}

\begin{figure}[!t]
\centering
\includegraphics[width=2.4in]{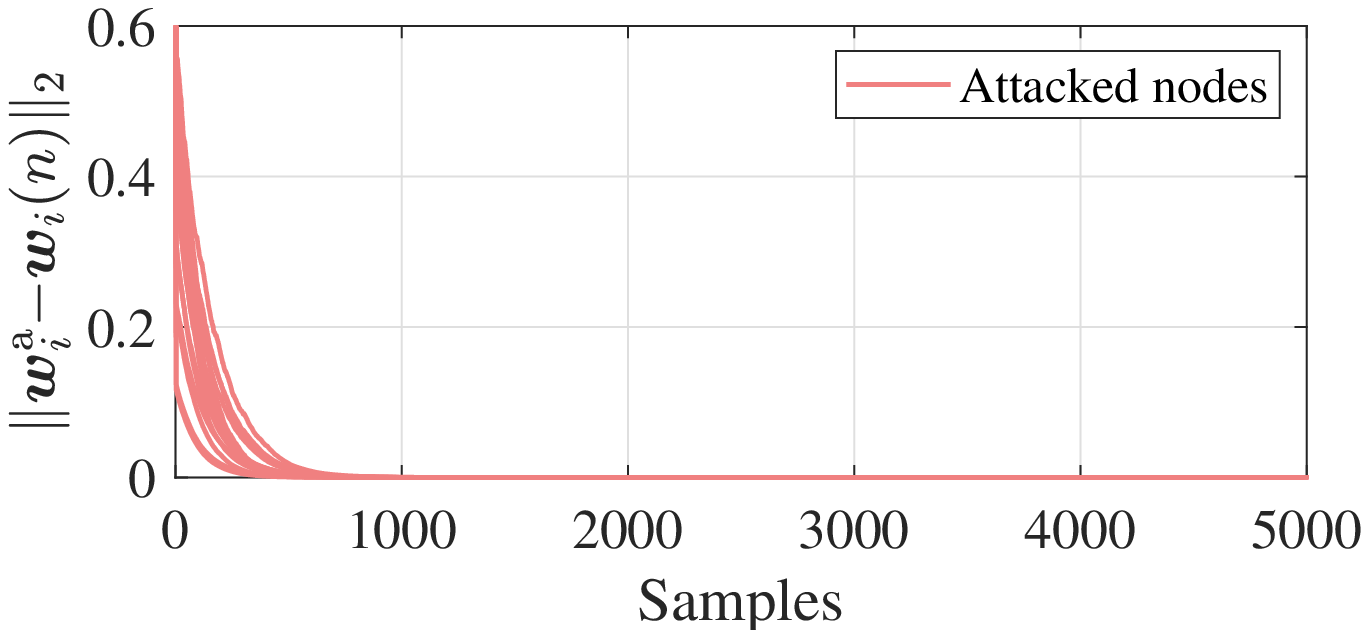}
\caption{Evolutions of distance between the attacked nodes and $\bm w_i^\mathrm{a}$.}
\label{fig:attacked_nodes_location}
\end{figure}

As previously mentioned, the DLMG algorithm with cooperation is not resilient and can be influenced by the Byzantine nodes. The step size $\mu_i = 0.02$ and the forgetting factor $\nu_i = 0.01$ of the DLMG algorithm have been taken uniformly across the normal nodes. The network topology at the end of the simulation with 5000 iterations performing the DLMG algorithm is shown in Fig. \ref{fig:end_network}. We observe that all light coral nodes, which previously belong to the normal nodes, are only connected with the Byzantine nodes illustrating the influence of malicious attacks. In addition, only the nodes sensing the identical target cluster together, which shows that blue nodes are connected with blue ones, and so do green nodes. Fig. \ref{fig:w1w2_location_without} shows the localization estimates for the $X$-coordinate $w_{i,1}(n)$ and the $Y$-coordinate $w_{i,2}(n)$ across all nodes during iterations. Fig. \ref{fig:attacked_nodes_location} shows the evolutions of distance between the attacked nodes and the malicious target $\bm w_i^\mathrm{a}$. It shows that all neighbors of Byzantine nodes reach the malicious target during a finite time, whereas all the other normal nodes achieve their desired targets, and naturally cluster two groups.

\begin{figure}[!t]
\centering
\includegraphics[width=1.6in]{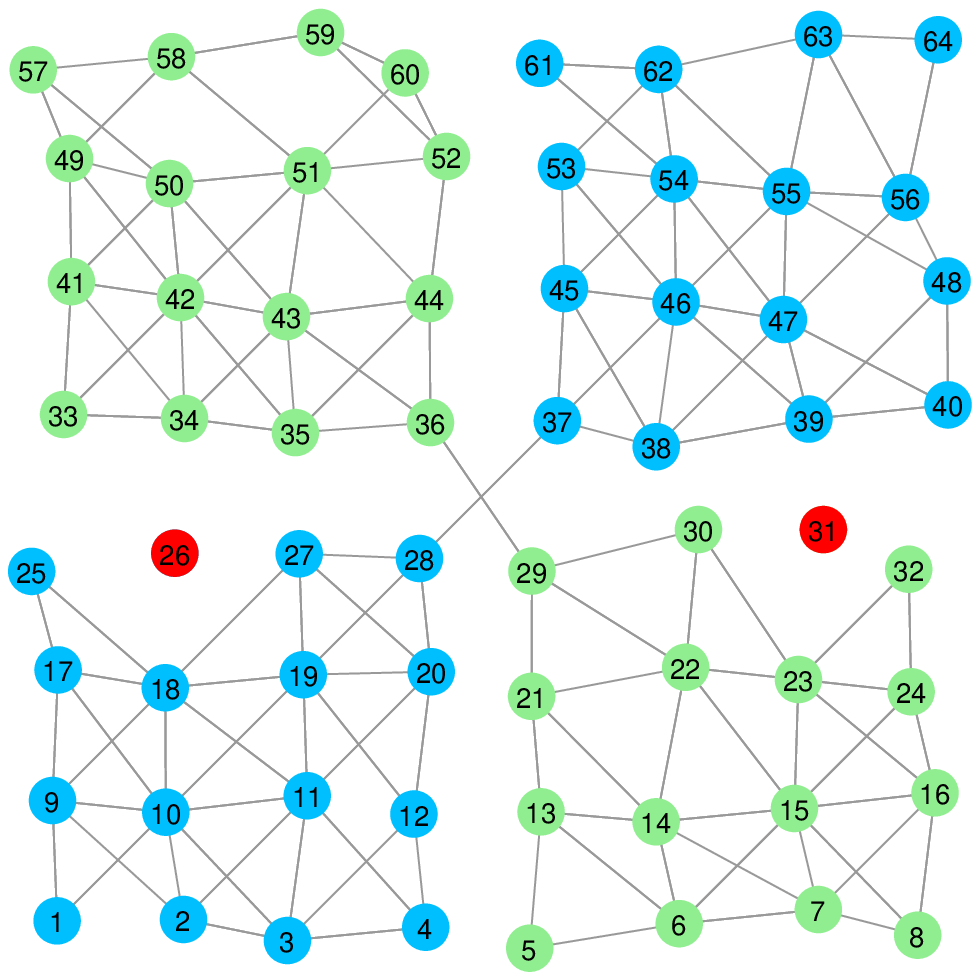}
\caption{Final network topology by executing RDLMG.}
\label{fig:end_network_resilient}
\vspace{-10pt}
\end{figure}

\begin{figure}[!t]
\centering
\includegraphics[width=2.5in]{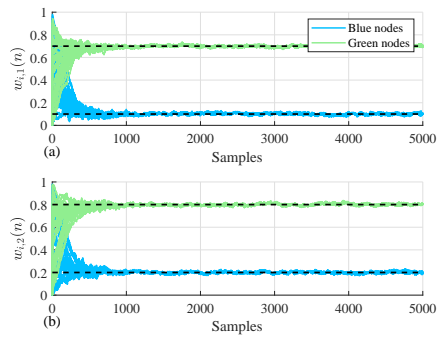}
\caption{Estimates of localization by executing RDLMG. (a) $X$-coordinate. (b) $Y$-coordinate.}
\label{fig:w1w2_location}
\vspace{-10pt}
\end{figure}

Since there is at most one Byzantine node among the neighbors of a normal node, in which each node discards the extreme value information of cost contributions received from its neighbors, it should be sufficient to guarantee all normal nodes to converge to their desired targets with resilience. After executing the RDLMG algorithm, Fig. \ref{fig:end_network_resilient} indicates that this is indeed the case with cutting all edges between normal nodes and Byzantine nodes, which shows that all normal nodes are divided into two connected sub-networks and Byzantine nodes are isolated. Fig. \ref{fig:w1w2_location} shows the localization estimates for the $X$-coordinate and the $Y$-coordinate across all normal nodes during iterations, which illustrates all normal nodes, whether they are neighbors of Byzantine nodes or not, will converge to their desired targets.

\begin{figure}[!t]
\centering
\includegraphics[width=2.5in]{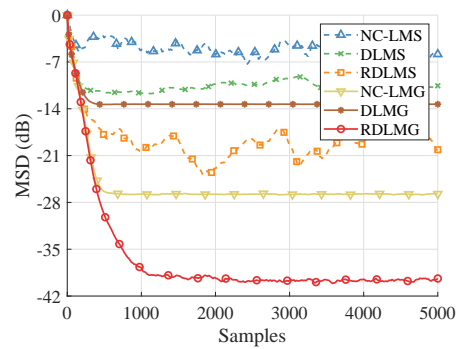}
\caption{Networked MSD evolutions of diffusion algorithms in the presence of CG noises and Byzantine attacks, with $p_i = 0.01$, $\sigma_{g_i}^2 = 10^4\sigma_{v_i}^2$ and $\mu_i^\mathrm{a}=0.01$.}
\label{fig:comparison_location}
\vspace{-10pt}
\end{figure}

We compare the performance of the proposed RDLMG algorithm with that of the LMS-based diffusion algorithms, NC-LMG and DLMG algorithms, where the networked MSD for all normal nodes is utilized as the performance metric. For a fair comparison, the parameters of various algorithms are set to guarantee the same initial evolutions. All results are obtained by averaging over independent runs. The networked MSD evolutions of diffusion algorithms in the presence of CG noises and Byzantine attacks are illustrated in Fig. \ref{fig:comparison_location}. It is worth noting that all LMS-based diffusion algorithms have experienced severe divergence under impulsive noises, even in the RDLMS algorithm. That is mainly because the MSE cost can not reduce the sensitivity to large outliers and LMS-based diffusion algorithms are not robust.
\begin{figure}[!t]
\centering
\includegraphics[width=2.5in]{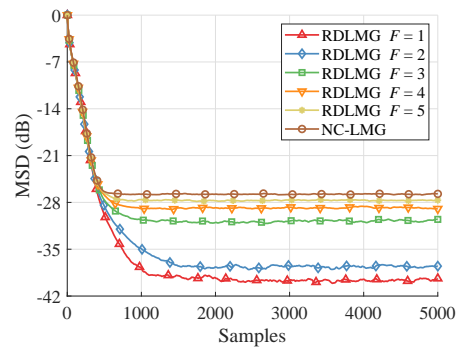}
\caption{Comparison of the networked MSD of NC-LMG and RDLMG with the selection of $F$.}
\label{fig:comparison_F_local}
\vspace{-10pt}
\end{figure}

Obviously, the DLMG strategy is also severely divergent and difficult to apply to cyber attacks environments. We note that if each node performs its own individual LMG algorithm, i.e., the NC-LMG algorithm, then the nodes do not cooperate with each other at all and Byzantine nodes can not influence the estimate of any nodes. So the multi-task network without cooperation is ``resilient'' in this sense. However, the networked MSD level of NC-LMG is quite high and its performance has seriously degraded. Thanks to the cooperation among nodes, RDLMG will possess much better performance than the NC-LMG algorithm. From Fig. \ref{fig:comparison_location}, we can see that the proposed RDLMG algorithm guarantees all normal nodes to converge to the desired targets in the presence of CG noises and Byzantine attacks, and also has lower MSD level as compared to the non-cooperative strategy. As stated in Remark 3, we can see from Fig. \ref{fig:comparison_F_local} that the selection of $F$ does not affect the convergence, but it influences the steady-state performance. As the increase of $F$, the steady-state networked MSD level also increases. In the ``worst-case'', normal nodes discard all the information from their neighbors, whose algorithm becomes NC-LMG and incurs the severe performance deterioration.

\begin{figure}[!t]
\centering
\includegraphics[width=2.5in]{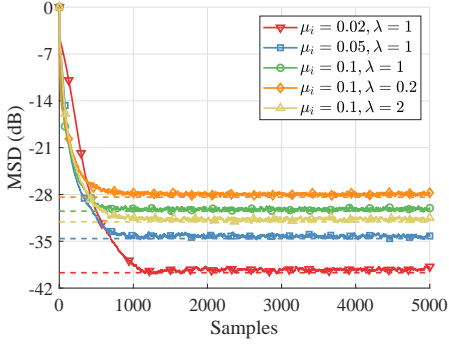}
\caption{Networked MSD evolutions of RDLMG with different $\mu_i$ and $\lambda$ in the presence of CG noises and Byzantine attacks, with $p_i = 0.01$, $\sigma_{g_i}^2 = 10^4\sigma_{v_i}^2$ and $\mu_i^\mathrm{a}=0.01$, where the dashed lines indicate the theoretical steady-state networked MSD.}
\label{fig:theory_values}
\vspace{-10pt}
\end{figure}

\begin{figure}[!t]
\centering
\includegraphics[width=2.5in]{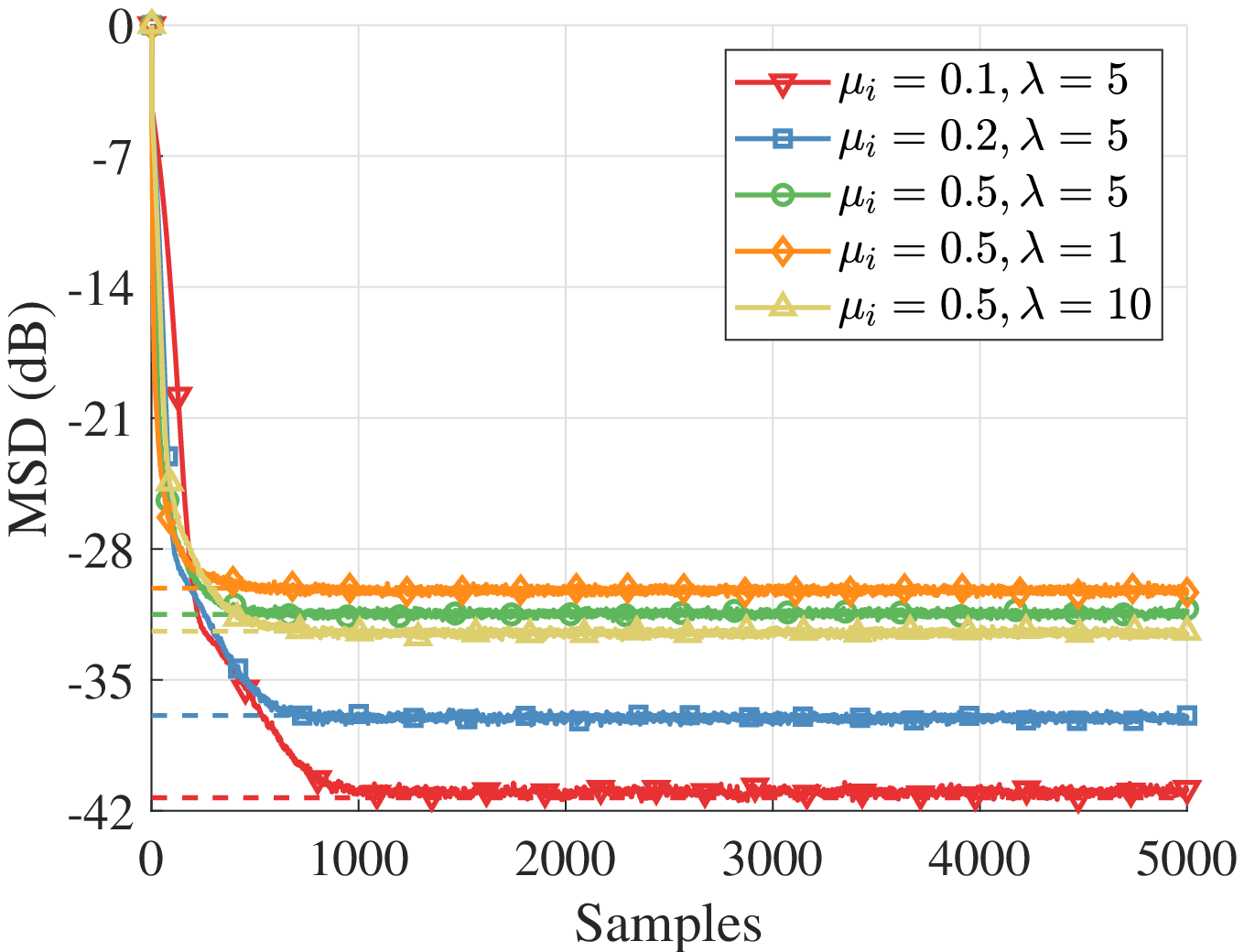}
\caption{Networked MSD evolutions of RDLMG with different $\mu_i$ and $\lambda$ in the presence of Gaussian noises and Byzantine attacks, with $\mathrm{SNR}=20$dB and $\mu_i^\mathrm{a}=0.01$, where the dashed lines indicate the theoretical steady-state networked MSD.}
\label{fig:theory_values_Gaussian}
\vspace{-10pt}
\end{figure}

Fig. \ref{fig:theory_values} investigates the analytical results of the RDLMG algorithm with different step sizes $\mu_i$ and scalar parameters $\lambda$, where the dashed lines indicate the theoretical steady-state networked MSD. The theoretical steady-state MSD for all normal nodes over the multi-task network is calculated as \eqref{eq:MSD_inf}. The simulated results of RDLMG are obtained by averaging over 100 independent runs. It is clear that the theoretical and simulated steady-state MSD match very well. We can also find that if $\mu_i$ is smaller, the convergence will be slower but it reduces the steady-state MSD level. A smaller $\lambda$ will lead to faster convergence and increased steady-state error. Fig. \ref{fig:theory_values_Gaussian} illustrates the networked MSD evolutions of RDLMG in the case of Gaussian noise with a signal-to-noise ratio (SNR) of 20dB. It shows that the RDLMG algorithm performs well and there is a good match between the theoretical and simulated steady-state MSD in the Gaussian interference scenario.

\begin{figure}[!t]
\centering
\includegraphics[width=2.5in]{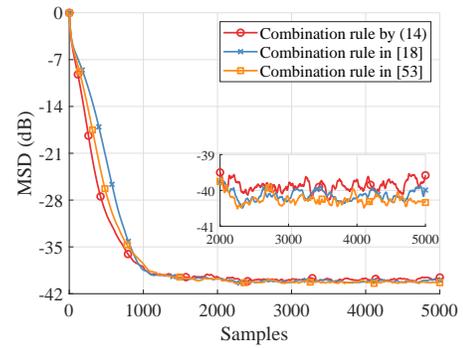}
\caption{Networked MSD evolutions of RDLMG with different combination rules in the presence of CG noises and Byzantine attacks, with $p_i = 0.01$, $\sigma_{g_i}^2 = 10^4\sigma_{v_i}^2$ and $\mu_i^\mathrm{a}=0.01$.}
\label{fig:combination_rules}
\vspace{-10pt}
\end{figure}
Fig. \ref{fig:combination_rules} illustrates the network MSD of RDLMG with the adaptive combination rule used by \eqref{eq:update_weights} and its variations mentioned in \cite{Chen2015Diffusion} and \cite{Nassif2016Diffusion}. It appears that all these combination rules are useful and they have the comparable performances. That is because each node discards the extreme value information of cost contributions received from its neighboring nodes via the removal set $\mathcal{R}_i(n+1)$. Therefore, the influence of different adaptive combination rules would be slight in such multi-task scenario.

\begin{figure}[!t]
\centering
\includegraphics[width=1.6in]{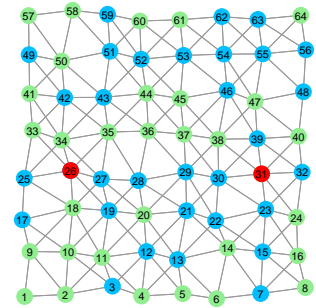}
\caption{Initial network topology with scattered nodes.}
\label{fig:initial_network_scatter}
\vspace{-10pt}
\end{figure}

\begin{figure}[!t]
\centering
\includegraphics[width=1.6in]{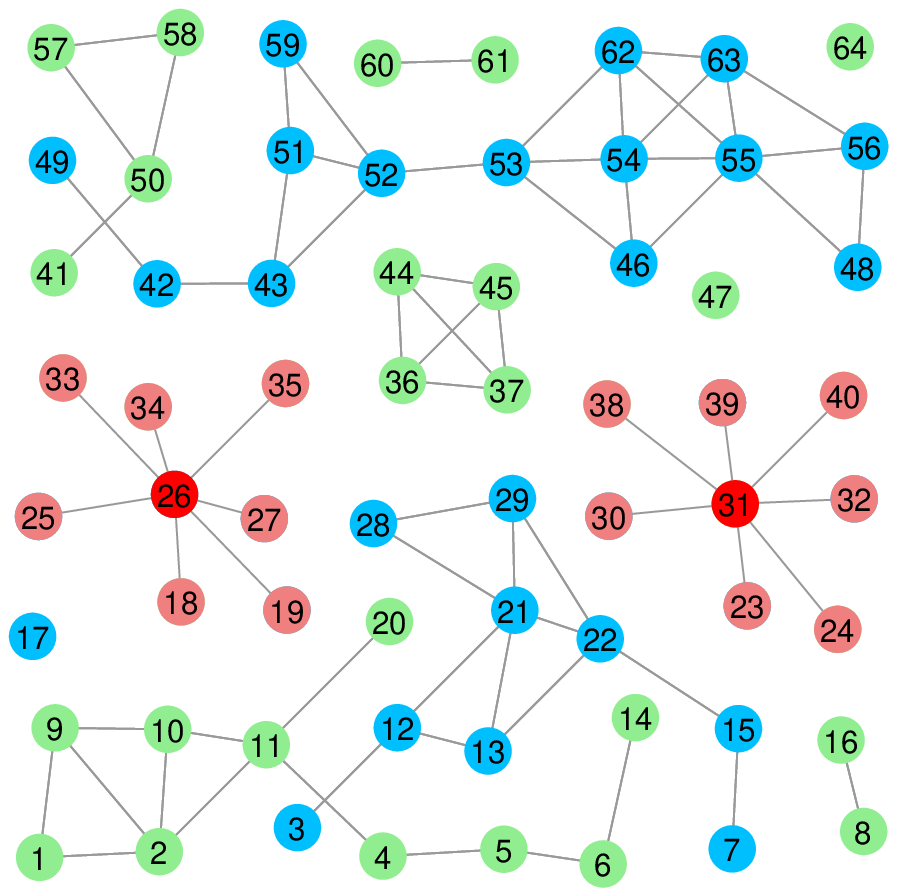}
\caption{Final network topology without resilience by executing DLMG.}
\label{fig:end_network_scatter}
\vspace{-10pt}
\end{figure}

\begin{figure}[!t]
\centering
\includegraphics[width=1.6in]{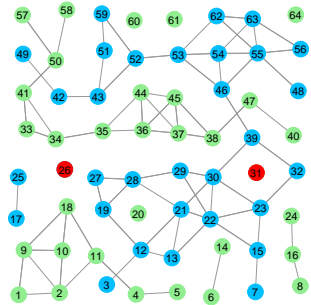}
\caption{Final network topology with resilience by executing RDLMG.}
\label{end_network_resilient_scatter}
\vspace{-10pt}
\end{figure}

Consider a connected network with scattered 64 nodes as shown in Fig. \ref{fig:initial_network_scatter}, with the same simulation configuration as previously mentioned. The network topology at the end of the iterations by executing DLMG is shown in Fig. \ref{fig:end_network_scatter}. We observe that the nodes previously belonging to the normal nodes are only connected with the Byzantine nodes illustrating the influence of malicious attacks. After executing RDLMG, Fig. \ref{end_network_resilient_scatter} shows that all links between normal nodes and the Byzantine nodes are removed, and Byzantine nodes are isolated. The normal node without neighbors will perform its own learning strategy. The resilience as we expected for the RDLMG algorithm has also been verified even though the nodes are scattered.


\subsection{Multi-Task Spectrum Sensing}

We have also evaluated the proposed RDLMG algorithm in multi-task spectrum sensing, which aims to estimate the spectrum of the transmitted signal in a connected network. Let $\Phi(e^{\mathrm{j}f})$ denote the power spectral density (PSD) of the transmitted signal at each frequency $f$ as
\begin{align}
\Phi(e^{\mathrm{j}f}) = \sum_{m=1}^M b_m(e^{\mathrm{j}f})w_m^\mathrm{o} = \bm b^\mathrm{T}(e^{\mathrm{j}f}) \bm w^\mathrm{o}
\end{align}
where $\bm b(e^{\mathrm{j}f})=[b_1(e^{\mathrm{j}f}),\dots,b_M(e^{\mathrm{j}f})]^\mathrm{T}$ represents the basis function vector evaluated at normalized frequency $f$ with length $M$, and $\bm w^\mathrm{o} = [w_1^\mathrm{o},\dots,w_M^\mathrm{o}]^\mathrm{T}$ is the weight vector representing the transmitted power over each basis. If $M$ is large enough, then the basis expansion can well approximate the transmitted spectrum and rectangular functions are commonly chosen for the set of basis $\{b_m(e^{\mathrm{j}f})\}_{m=1}^M$ \cite{DiLorenzo2013Distributed,Miller2016Distributed}.

Let $H_i(e^{\mathrm{j}f},n)$ be the channel transfer function between the transfer node and the receiver node $i$ at time $n$, and then the PSD received by node $i$ can be formulated as
\begin{align}
\Phi_i(e^{\mathrm{j}f},n) &= \sum_{m=1}^M |H_i(e^{\mathrm{j}f},n)|^2 b_m(e^{\mathrm{j}f})w_{i,m}^\mathrm{o} + \sigma_{r_i}^2 \nonumber\\
&= \bm b_i^\mathrm{T}(e^{\mathrm{j}f},n) \bm w_i^\mathrm{o} + \sigma_{r_i}^2
\end{align}
where $\bm b_i(e^{\mathrm{j}f},n)=|H_i(e^{\mathrm{j}f},n)|^2 [b_1(e^{\mathrm{j}f}),\dots,b_M(e^{\mathrm{j}f})]^\mathrm{T}$, and the weight vector $\bm w_i^\mathrm{o} = [w_{i,1}^\mathrm{o},\dots,w_{i,M}^\mathrm{o}]^\mathrm{T}$ needs to be estimated, and $\sigma_{r_i}^2$ represents the receiver noise power at node $i$. At time $n$, each node $i$ measures the received PSD within the normalized frequency range $[f_{\min},f_{\max}]=[0,1]$ over $M_f$ frequency samples $f_\iota = f_{\min}:(f_{\max}-f_{\min})/M_f:f_{\max}, \iota=1,\dots,M_f$. Therefore, the noisy desired signal at frequency $f_\iota$ is given by
\begin{align}\label{eq:noisy_signal}
d_{i,\iota}(n) = \bm b_i^\mathrm{T}(e^{\mathrm{j}f_\iota},n) \bm w_i^\mathrm{o} + \sigma_{r_i}^2 + \eta_{i,\iota}(n)
\end{align}
where $\eta_{i,\iota}(n)$ denotes the additive measurement noise at frequency $f_\iota$.

Using an energy estimator over an idle band, the receiver noise power $\sigma_{r_i}^2$ can be accurately estimated, and thus subtracted form \eqref{eq:noisy_signal} before spectrum sensing \cite{Miller2016Distributed}. Collecting the noisy desired signal over $M_f$ frequencies, a linear model for spectrum sensing at every node $i$ can be obtained by
\begin{align}\label{eq:PSD}
\bm d_i(n) = \bm B_i(n) \bm w_i^\mathrm{o} + \bm \eta_i(n)
\end{align}
where we denote $\bm d_i(n) = [d_{i,1}(n),\dots,d_{i,M_f}(n)]^\mathrm{T} \in \mathbb{R}^{M_f}$, $\bm B_i(n)=[\bm b_i(e^{\mathrm{j}f_1},n),\dots,\bm b_i^\mathrm{T}(e^{\mathrm{j}f_{M_f}},n)] \in \mathbb{R}^{M_f \times M}$, and $\bm \eta_i(n) = [\eta_{i,1}(n),\dots,\eta_{i,M_f}(n)]^\mathrm{T} \in \mathbb{R}^{M_f}$.

\begin{figure}[!t]
\centering
\includegraphics[width=1.6in]{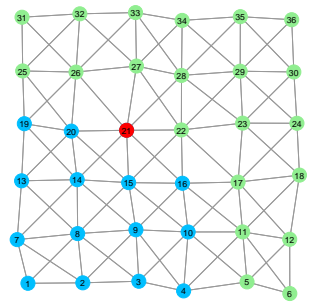}
\caption{Initial multi-task network topology.}
\label{fig:initial_network_sensing}
\vspace{-10pt}
\end{figure}
Consider a connected network with 36 nodes for the clustered multi-task scenario as illustrated in Fig. \ref{fig:initial_network_sensing}, which comprises of 35 normal nodes divided into two groups with blue and green nodes, and contains one Byzantine node in red. Each normal node can be modeled as \eqref{eq:PSD} for spectrum sensing, in which we utilize $M=50$ non-overlapping rectangular basis functions with amplitude equal to one to well approximate the transmitted spectrum. The nodes scan $M_f=100$ frequencies over the normalized frequency axis. We assume that $w_i^\mathrm{o}$ has 8 non-zero elements, which means that the corresponding spectrum is transmitted over 8 basis functions, and the power transmitted over each basis function is set to 0.7. The multi-task spectrum sensing aims to estimate $w_i^\mathrm{o}$ for each node, where the 8 non-zero elements of spectrum coefficients $w_i^\mathrm{o}$ for blue and green nodes only locate in different positions.
\begin{figure}[!t]
\centering
\includegraphics[width=1.6in]{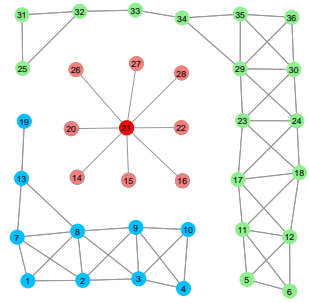}
\caption{Final network topology by executing DLMG.}
\label{fig:end_network_sensing}
\vspace{-10pt}
\end{figure}
\begin{figure}[!t]
\centering
\includegraphics[width=2.5in]{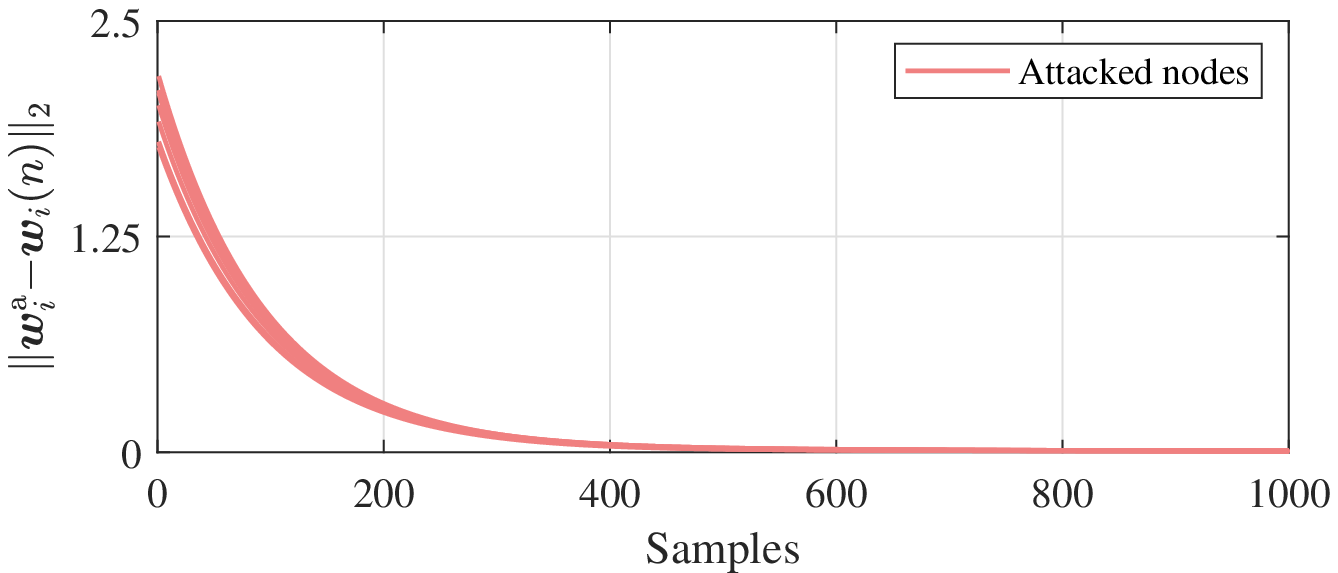}
\caption{Evolutions of distance between the attacked nodes and $\bm w_i^\mathrm{a}$.}
\label{fig:attacked_nodes_sensing}
\vspace{-10pt}
\end{figure}
\begin{figure}[!t]
\centering
\includegraphics[width=1.6in]{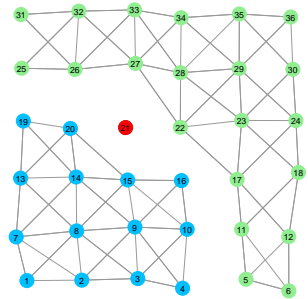}
\caption{Final network topology by executing RDLMG.}
\label{fig:end_network_resilient_sensing}
\vspace{-10pt}
\end{figure}

The additive noise $\eta_{i,\iota}(n)$ is assumed to be a combination of the background noise $v_{i,\iota}(n)$ and the impulsive noise $\omega_{i,\iota}(n)$, i.e., $\eta_{i,\iota}(n) = v_{i,\iota}(n) + \omega_{i,\iota}(n)$. The background noise $v_{i,\iota}(n)$ is extracted from a zero-mean Gaussian sequence with a variance depending on the interference level as $\sigma_{v_{i,\iota}}^2 = \sigma_{r_i}^2[\sigma_{r_i}^2+2\Phi_i(e^{\mathrm{j}f_\iota})]/10$ at frequency $f_\iota$, where the receiver noise power is set to $\sigma_{r_i}^2 = 0.1$. We consider that the $\alpha$-stable distribution is used to model the impulsive feature, whose characteristic function is expressed as $\phi(t) = e^{-\gamma^\alpha|t|^\alpha}$, where the characteristic parameter is $0< \alpha \le 2$ with the smaller $\alpha$ leading to more outliers, and the dispersion parameter is $\gamma>0$ \cite{Yu2019Robust}. The impulsive noise $\omega_{i,\iota}(n)$ is drawn from an $\alpha$-stable process with $\alpha =1.2$ and $\gamma = 0.08$.

The Byzantine node aims to drive its neighboring normal nodes to estimate the malicious spectrum coefficient as $\bm w_i^\mathrm{a} = 0.5 \bm 1_M$, which means that all malicious powers are set to 0.5, and the attack parameter is selected as $\mu_i^\mathrm{a}=0.01$.

The step size $\mu_i = 0.2$ and the forgetting factor $\nu_i = 0.001$ of the DLMG algorithm have been taken uniformly across the normal nodes. The network topology at the end of the iterations by executing the DLMG algorithm is shown in Fig. \ref{fig:end_network_sensing}. Fig. \ref{fig:attacked_nodes_sensing} shows the evolutions of distance between the attacked nodes and the malicious spectrum coefficient $\bm w_i^\mathrm{a}$. We can clearly see that all neighbors influenced by the Byzantine node will converge to the malicious target. After executing the RDLMG algorithm, Fig. \ref{fig:end_network_resilient_sensing} shows that all normal nodes cluster two sub-networks and all links between normal nodes and the Byzantine node are removed, which demonstrates the resilience as we expected.
\begin{figure}[!t]
\centering
\includegraphics[width=2.5in]{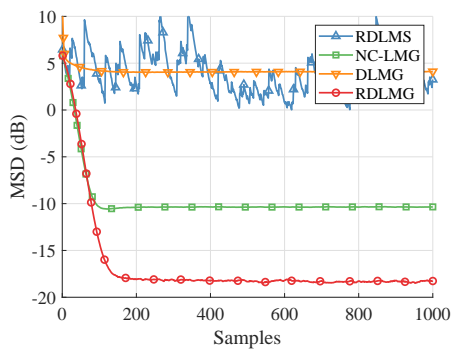}
\caption{Networked MSD evolutions of diffusion algorithms in the presence of $\alpha$-stable noises and Byzantine attack, with $\alpha = 1.2$, $\gamma = 0.08$ and $\mu_i^\mathrm{a}=0.01$.}
\label{fig:comparison_sensing}
\vspace{-10pt}
\end{figure}
\begin{figure}[!t]
\centering
\includegraphics[width=2.5in]{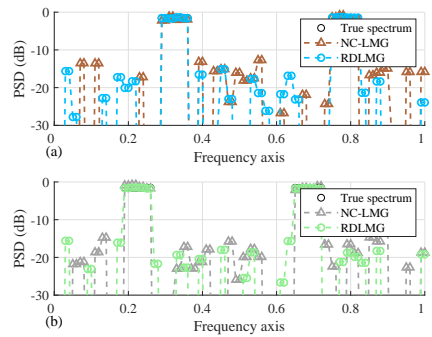}
\caption{Spectrum estimation of distributed algorithms. (a) A node depicted in blue task. (b) A node depicted in green task.}
\label{fig:w1w2}
\vspace{-10pt}
\end{figure}

To assess the performance of diffusion algorithms, the networked MSD for all normal nodes in the presence of $\alpha$-stable noises and Byzantine attack are reported in Fig. \ref{fig:comparison_sensing}. All results are obtained by averaging over independent runs. It is observed that the RDLMS and DLMG algorithms have severe performance divergence in this situation. It is obvious that the proposed RDLMG algorithm with cooperation guarantees convergence to the true spectrum and also results in lower MSD level in comparison with NC-LMG.

We evaluate the results of multi-task spectrum estimation in the impulsive noises and cyber attacks environments, carried out by the convergent RDLMG and NC-LMG algorithms. The muli-task spectrum estimation of distributed algorithms in accordance with PSD is exhibited in Fig. \ref{fig:w1w2}. As we can notice, the RDLMG algorithm is able to make the accurate spectrum estimation and has lower side lobes in terms of PSD. It is also proved that the proposed RDLMG algorithm enables to fit much better the transmitted spectrum due to the inter-node cooperation than that of the NC-LMG algorithm.

\section{Conclusion}

In this paper, we have proposed a novel robust resilient algorithm over clustered multi-task networks in the presence of impulsive interferences and Byzantine attacks. The proposed RDLMG algorithm relies on the GM estimator and the MSR strategies, which are able to counteract the adverse effects of impulsive noises and Byzantine nodes. Theoretical performance analyses of the RDLMG algorithm in the mean and mean-square senses were developed. Numerical results validate the performance of RDLMG and the theoretical results. We have also evaluated the proposed RDLMG algorithm by applying it to the problems of multi-target localization and multi-task spectrum sensing, which all demonstrate its superior performance to some existing distributed algorithms.

\bibliographystyle{IEEEtran}
\bibliography{IEEEref}

\end{document}